\PassOptionsToPackage{dvipsnames,table}{xcolor} %
\RequirePackage{fix-cm} %
\documentclass[11pt,a4paper]{newtemplate}

\usepackage{amsmath,amsfonts,bm}

\def\eqref#1{equation~\ref{#1}}

\def\1{\bm{1}}

\DeclareMathAlphabet{\mathsfit}{\encodingdefault}{\sfdefault}{m}{sl}
\SetMathAlphabet{\mathsfit}{bold}{\encodingdefault}{\sfdefault}{bx}{n}

\usepackage{natbib}
\usepackage{float}
\usepackage{twemojis}

\usepackage[percent]{overpic}
\usepackage{alltt}

\definecolor{myblue}{HTML}{3f55b5}
\definecolor{myorange}{HTML}{ec6b14}
\hypersetup{colorlinks=true, urlcolor=myorange, linkcolor=red, citecolor=myblue}
\newcommand{\wordmark}{BrickBench}
\usepackage{cleveref}

\crefname{figure}{Fig.}{Figs.}
\crefname{table}{Tab.}{Tabs.}
\crefname{equation}{Eq.}{Eqs.}
\crefname{section}{Sec.}{Secs.}
\crefname{subsection}{Sec.}{Secs.}
\crefname{subsubsection}{Sec.}{Secs.}
\Crefname{figure}{Fig.}{Figs.}
\Crefname{table}{Tab.}{Tabs.}
\Crefname{equation}{Eq.}{Eqs.}
\Crefname{section}{Sec.}{Secs.}
\Crefname{subsection}{Sec.}{Secs.}
\Crefname{subsubsection}{Sec.}{Secs.}

\usepackage{subcaption}
\usepackage{tikz}
\ifdefined\AddToHook
\makeatletter
\AddToHook{begindocument}[brk/quiet]{\let\brk@check@eq\@check@eq \def\@check@eq#1#2{}}
\AddToHook{begindocument}[brk/restore]{\let\@check@eq\brk@check@eq}
\DeclareHookRule{begindocument}{brk/quiet}{before}{microtype}
\DeclareHookRule{begindocument}{brk/restore}{after}{microtype}
\makeatother
\fi
\DeclareMathSizes{5.2}{5.2}{5}{5}

\newcommand{\PillOuterFrac}{0.98} %
\newdimen\PillXPad   \PillXPad=4.0pt
\newdimen\PillYPad   \PillYPad=1.4pt
\newdimen\PillRule   \PillRule=0.25pt
\newdimen\PillRadius \PillRadius=3.0pt %

\tikzset{
  cvprPill/.style={
    rounded corners=\the\PillRadius,
    draw=black!20,
    fill=black!2,
    line width=\the\PillRule,
    inner xsep=\the\PillXPad,
    inner ysep=\the\PillYPad,
    outer sep=0pt,
    line cap=round,
    line join=round
  }
}

\newsavebox{\capline}

\DeclareCaptionFormat{cvprpill}{%
  \begingroup
  \sbox{\capline}{\strut #1\hspace{0.35em}#3}%

  \dimen0=\dimexpr\PillOuterFrac\hsize-\PillXPad-\PillXPad-\PillRule-\PillRule\relax

  \noindent\hbox to\hsize{\hss
    \tikz[baseline=(cap.base)]\node[cvprPill] (cap) {%
      \microtypesetup{protrusion=false}%
      \ifdim\wd\capline<\dimen0
        \strut\noindent\ignorespaces#1\hspace{0.35em}#3\unskip
      \else
        \begin{minipage}{\dimen0}
          \microtypesetup{protrusion=false}%
          \setlength{\parindent}{0pt}\setlength{\parskip}{0pt}%
          \leftskip=0pt \rightskip=0pt
          \parfillskip=0pt plus 1fil %
          \spaceskip=0pt \xspaceskip=0pt
          \hyphenpenalty=10000 \exhyphenpenalty=10000
          \tolerance=2000 \emergencystretch=1em
          \noindent\ignorespaces#1\hspace{0.35em}#3\unskip
        \end{minipage}
      \fi
    };%
  \hss}%
  \endgroup
}

\definecolor{tsrInk}{HTML}{202124}
\definecolor{tsrModel}{HTML}{E5DBFF}
\definecolor{tsrSet}{HTML}{FFE8CC}
\definecolor{tsrAlt}{HTML}{D3F9D8}
\definecolor{tsrChip}{HTML}{F1F3F5}
\definecolor{bnInk}{HTML}{202124}
\definecolor{bnFill}{HTML}{FFDEEB}
\newcommand{\tsrmono}{\fontencoding{T1}\fontfamily{fvm}\fontseries{b}\selectfont}
\tikzset{
  tsrbox/.style={draw=tsrInk, line width=0.9pt, rounded corners=4pt, inner sep=0pt, outer sep=0pt},
}

\newcommand{\tsrpromptfont}{\fontsize{8}{9.4}\selectfont}
\newcommand{\teaserrow}[4]{%
  \noindent\includegraphics[width=\textwidth]{fig/images/#4}\par
  \vspace{1.5pt}\nointerlineskip\noindent
  \tikz\node[tsrbox, line width=0.6pt, fill=#2, text width=\dimexpr\textwidth-5pt\relax, inner xsep=2pt,
             inner ysep=1.5pt, align=left, font=\sffamily\tsrpromptfont]
    {{\tsrmono\tsrpromptfont #1}\enspace #3};%
  \par\vspace{1.5pt}\nointerlineskip}

\newlength{\tsrcap}\newlength{\tsrlogo}
\newcommand{\teasercol}[3]{%
  \makebox[\dimexpr\textwidth/5\relax][c]{\makebox[80pt][l]{%
    \sffamily\fontsize{8}{9}\selectfont
    \settoheight{\tsrcap}{H}\setlength{\tsrlogo}{12pt}%
    \raisebox{\dimexpr0.5\tsrcap-0.5\tsrlogo\relax}{\includegraphics[height=\tsrlogo]{fig/icons/#1}}%
    \hspace{3pt}#2 #3}}}

\title{\textls[10]{\textbf{\wordmark}: Evaluating Agentic Brick Design}}

\renewcommand\Affilfont{\normalfont\bfseries\small\centering}
\makeatletter\renewcommand\AB@affilsepx{\protect\\\protect\Affilfont}\makeatother
\author{Peter Kulits$^{1,2}$ \quad Yiqing Xu$^{1}$ \quad R.~Kenny Jones$^{1}$ \quad Cordelia Schmid$^{3}$ \quad Jiajun Wu$^{1}$ \\
$^{1}$Stanford University \quad $^{2}$Max Planck Institute for Intelligent Systems \quad $^{3}$Inria
}

\fancypagestyle{firststyle}{
  \fancyhf{}
  \fancyhead[L]{{\small \textit{\the\month-\the\day-\the\year}}}
  
}

\teaser{\begingroup\setlength{\intextsep}{3pt}\setlength{\abovecaptionskip}{4pt}
\begin{figure}[H]
  \centering
  \teaserrow{Model}{tsrModel}{A pelican riding a bicycle.}{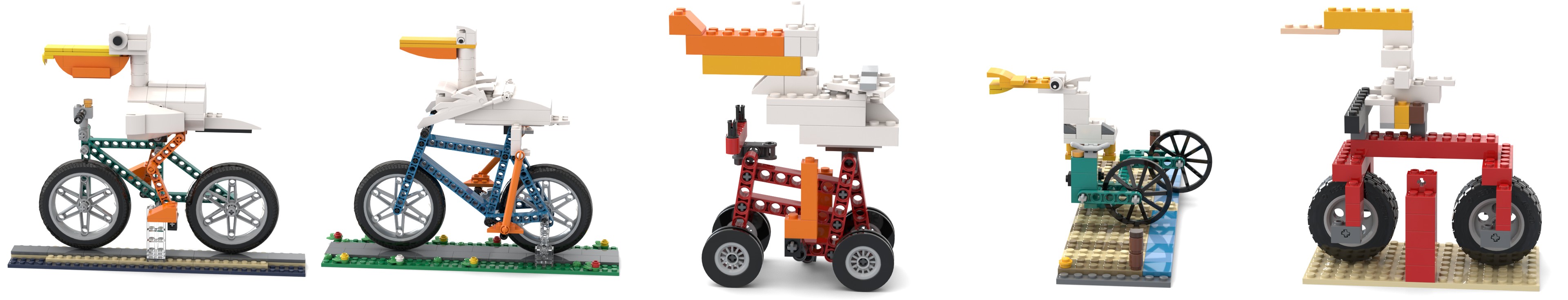}
  \teaserrow{Set}{tsrSet}{A brown dog with floppy black ears and a small brown hat sits at a table in a burning room. A white coffee mug rests in front of it. Flames climb the curtains and surround the chair, but the dog sits upright with its paws resting calmly on the tabletop. The front of the room is open to reveal the scene.}{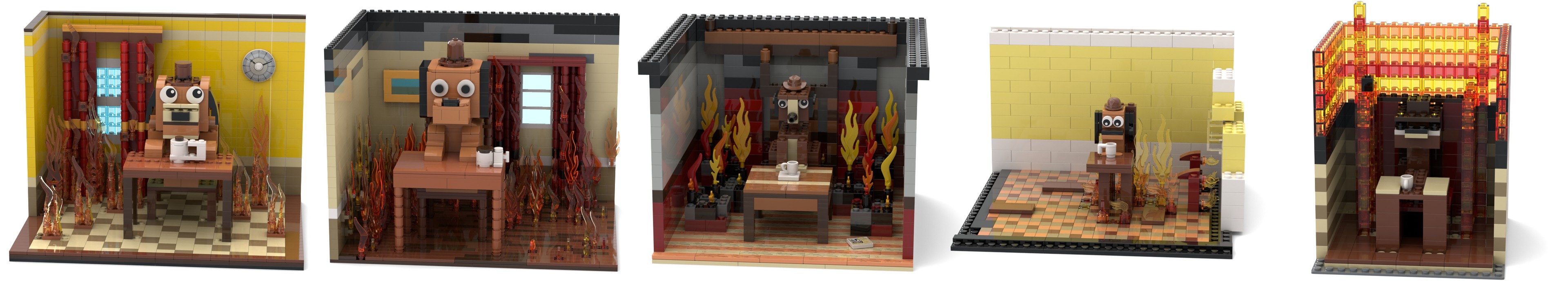}
  \teaserrow{Alt-Build}{tsrAlt}{A cowboy in a wide-brimmed hat sits on a saddled horse beside a cactus with two raised arms. He holds a looped lasso above his head. The horse's front legs stand on a low rock, while its rear legs remain on the sandy ground.}{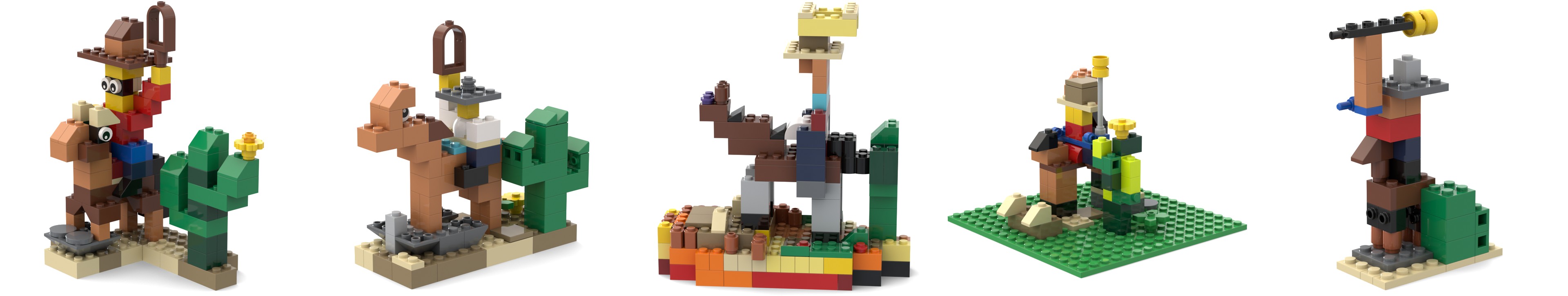}
  \vspace{2pt}\noindent%
  \teasercol{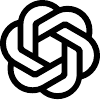}{GPT-6}{Astra}%
  \teasercol{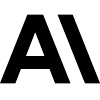}{Claude}{Opus 5.5}%
  \teasercol{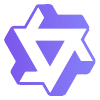}{Qwen 3.8}{Flash}%
  \teasercol{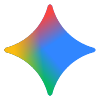}{Gemini 3.8}{Flash}%
  \teasercol{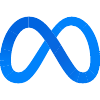}{Muse}{Spark 1.3}%
  \caption{We evaluate coding agents on their ability to design LEGO assemblies from a text prompt. We introduce three settings with varying constraints: \texttt{Model} ($\leq$400 parts), \texttt{Set} (400--4000 parts), and \texttt{Alt-Build} (the part inventory of set 10698).
  Each assembly above can be built.
  }
  \label{fig:teaser}
\end{figure}
\endgroup
}

\begin{document}

\begin{abstract}
We propose BrickBench, a benchmark for agentic text-conditioned LEGO-set design.
Given a prompt, an agent is tasked with producing an assembly that not only satisfies semantic and design criteria, but that can also be physically built.
To do so, it must select parts from a discrete library and reason jointly about local and global constraints.
We score validity, alignment, and design across three settings that vary in scale and part availability.
We provide BrickAgent, an environment for coding agents to construct, inspect, and validate their designs.
We find that leading agents largely satisfy verifiable physical and semantic requirements, but fall short of human designs.
We release our benchmark and environment at \url{http://www.brickben.ch}
\end{abstract}

\maketitle

\section{Introduction}\label{sec:introduction}
\setlength{\parskip}{4pt}

Designing a LEGO assembly is deceptively difficult.
A designer must select from thousands of parts, arrange them into a recognizable form, and ensure that the structure holds together.
Computational generative approaches to brick assembly have relied on specialized models and carefully engineered representations to handle these challenges~\citep{chung2021brick,Pun_2025_ICCV,Kulits_2026_CVPR}.
But coding agents offer a different approach: they can write programs, execute them, inspect the results, and revise.
This shift is already transforming software engineering~\citep{jimenez2024swebench}, graphics~\citep{gu2025blendergym}, and CAD~\citep{zhang2026benchcad}.
We find that a general coding agent, without task-specific training, substantially outperforms specialized LEGO-generation models on the BrickNet benchmark~\citep{Kulits_2026_CVPR}, approaching the semantic alignment of its reference assemblies (\cref{fig:bricknet}).

Yet does this mean agents have learned to design LEGO?
Existing evaluations leave two important challenges unexplored.
First, \emph{physical complexity:} BrickNet evaluates small objects of at most 100 parts, far from the hundreds or thousands of interconnected pieces and inventory constraints of a full LEGO set.
Second, and more fundamentally, \emph{design quality:} making an assembly that works is not the same as making an assembly worth building.
The appeal of LEGO lies not merely in reproducing an object, but in the creativity of its construction: clever part usage, shapes, proportions, and details.
Two builds can be equally valid and faithful to a description, yet differ dramatically in how well they are designed.
Unlike physical validity or semantic alignment with a prompt, design quality has no definitive pass/fail test.

We introduce \textbf{BrickBench}, a benchmark for agentic text-conditioned LEGO-set design (\cref{fig:teaser}).
BrickBench contains 300 prompts across three settings: \texttt{Model} (at most 400 parts), \texttt{Set} (400--4000 parts), and \texttt{Alt-Build} (a fixed inventory of parts).
To support complex construction, we introduce \textbf{BrickAgent}, an environment for agentic LEGO design including construction tools and interpretable validation of connectivity, collisions, and stability.
Rather than prescribing how to design an assembly, BrickAgent provides the building blocks, constraints, and feedback, leaving the agent to devise its own solution.
To evaluate what validity and semantic alignment miss, we measure \emph{design quality} through pairwise judgments, validated against human preferences.

We evaluate eleven frontier agents and two data-driven baselines.
With BrickAgent, leading agents produce valid assemblies for nearly every prompt, and the best satisfies approximately 95\% of semantic requirements.
Removing the environment sharply reduces physical validity, showing that reliable construction depends on grounded tools and feedback.
Design quality, however, separates agents far more strongly, and human raters identify human-designed assemblies in 323 of 360 comparisons.
These findings reveal a shift in the challenge of LEGO design: \emph{Agents are learning to build what works, but not yet what makes a great design.}

Our contributions are as follows.
\begin{enumerate}[leftmargin=2em,itemsep=0pt,parsep=0pt,partopsep=0pt,topsep=1pt]
  \item \textbf{BrickBench}, a benchmark of 300 prompts in three settings for agentic brick design, with metrics for physical validity (connections, collisions, and stability under simulation), semantic alignment (\texttt{VQA}, over yes/no questions decomposed from each prompt), and design quality (\texttt{ELO}, from pairwise judgments by a VLM validated against human raters).
  \item \textbf{BrickAgent}, an environment enabling coding agents to programmatically construct, inspect, and validate complex LEGO assemblies, with part search, placement by connectors, subassemblies, rendering, and a validator that names parts at fault. Without it, the share of valid assemblies falls from 100\% to 40\% for GPT-6 Astra and to under 1\% for GPT-5.6 Luna.
  \item \textbf{A systematic evaluation} of eleven frontier agents and two data-driven baselines, with human studies that validate the design metric and measure the gap to human design.
\end{enumerate}

\section{Related Work}\label{sec:related}

\noindent\textbf{Agentic Benchmarks.}\label{ssec:related-agentic}
Benchmarks for coding agents validate an agent's work by executing it.
SWE-bench~\citep{jimenez2024swebench} runs a repository's tests on a patch an agent supplied, and OSWorld~\citep{xie2024osworld} tracks state through execution.
SceneCraft~\citep{hu2024scenecraft} writes Blender scripts, and CAD-Assistant~\citep{mallis2025cadassistant} writes CAD programs.
Benchmarks then execute the program and score the result.
BlenderGym~\citep{gu2025blendergym} compares the result with a goal scene, and BenchCAD~\citep{zhang2026benchcad} executes CadQuery programs for industrial parts.
BuildArena~\citep{xia2026buildarena} scores machines assembled from modular parts by their simulated behavior, and PhyBlock~\citep{ma2025phyblock} scores block placements for plausibility in a physics simulator.
BrickBench executes the agent's program into an assembly of real LEGO parts and validates connections, collisions, and stability under simulation.

\noindent\textbf{LEGO and Evaluation.}\label{ssec:related-lego}
Prior work generates LEGO assemblies with learned models and uses LEGO tasks to evaluate others.
BrickGPT~\citep{Pun_2025_ICCV} fine-tunes a language model to place eight brick types on a $20{\times}20{\times}20$ grid from a caption.
BrickNet~\citep{Kulits_2026_CVPR}, predicts connections rather than positions and autoregressively generates structures through typed connectors.
LEGO-Maker~\citep{ge2025legomaker} produces building facades conditioned on an image.
Other work evaluates agents and multimodal models on LEGO tasks.
Break and Make~\citep{walsman2022break} gives an agent an LDraw model in a simulator to test whether it can inspect it, take it apart, and rebuild it from scratch.
LEGO-Puzzles~\citep{tang2026lego} presents multimodal models spatial questions about assembly steps.
BrickBench does both: general coding agents generate set-scale assemblies from a text prompt, while we score the assemblies on physical validity, prompt alignment, and design.

\section{Preliminaries}\label{sec:preliminaries}

\subsection{LDraw}\label{ssec:ldraw}
LDraw is a community-maintained part library and standard for digital LEGO modeling\footnote{\url{https://www.ldraw.org}}.
Each library part is intended to mirror a real-world counter-part.
Assemblies are represented as text files of instances defined by their part type, color, and 6D pose.
LDraw does not itself, however, determine whether an assembly holds together or could be built.

\subsection{Connectivity and Physical Validity}\label{ssec:validity}
BrickNet~\citep{Kulits_2026_CVPR} annotates each part in the LDraw library with typed connectors.
Each connector instance has a type, a subtype, a polarity, and a frame.
The subtype names the connector's geometry within its type and fixes what it mates with: a stud mates with a hole or a tube, and a bar with a clip.
The polarity distinguishes the two halves of a hinge, ball, or fixed joint, which must be opposite to connect.
The frame is the connector's position and orientation in the coordinates of its part, with one axis along the direction of mating.
Two part instances are connected when a compatible pair of their connectors is aligned up to tolerance.

There are five connector types.
A stud connector allows rotation about the connection axis, a hinge allows rotation and a possible flip, and an axle additionally allows a slide along its axis.
A ball allows free rotation, and a fixed-type connector allows no motion.

The connections between instances of an assembly form a graph over parts.
An assembly may have more than one component to represent a scene with multiple (disconnected) objects.
We define an assembly as stable if it remains at rest under gravity.
We treat connected components as rigid bodies of the assembly and simulate them as single rigid bodies of uniform density in PyBullet~\citep{coumans2021pybullet}.
We set a high coefficient of friction and no restitution, so that inter-component contacts neither slide nor bounce.
A component then fails by falling when it is not supported.
We consider an assembly stable if no point of any component moves more than three quarters the height of the stud of a brick, 3 LDraw Units (LDU).
Effects such as the holding force of stud connections and the play of axles are not modeled, so an assembly that balances as a rigid body passes even when a built one might sag or fall apart.
Evaluating this would require a structural simulation, with either a force limit at each connection or a finite-element model of the parts, which we discuss in \cref{sec:discussion}.
Two parts collide when their geometry overlaps by more than a tolerance.
An assembly with collisions cannot be built, and is determined invalid prior to simulation.

\subsection{Text-Conditioned Assembly}\label{ssec:text-assembly}
\providecommand{\tsrmono}{\fontencoding{T1}\fontfamily{fvm}\fontseries{b}\selectfont}

\newcommand{\bnrow}[2]{%
  \noindent\includegraphics[width=\linewidth]{fig/images/#1}\par
  \vspace{1.5pt}\nointerlineskip\noindent
  \tikz\node[draw=bnInk, line width=0.6pt, rounded corners=3pt, fill=bnFill, inner xsep=2pt, inner ysep=1.5pt,
             outer sep=0pt, text width=\dimexpr\linewidth-7pt\relax, align=left,
             font=\sffamily\fontsize{6}{7}\selectfont] {#2};%
  \par\vspace{1.5pt}\nointerlineskip}

\newcommand{\bncol}[1]{\makebox[0.2495\linewidth][c]{\sffamily\fontsize{7}{8}\selectfont #1}}

\newsavebox{\bnpanel}
\providecommand{\bnleftw}{0.467\textwidth}
\begin{figure}[t]
  \centering
  \captionsetup[subfigure]{skip=2pt}%
  \sbox{\bnpanel}{\begin{minipage}[b]{\bnleftw}
    \bnrow{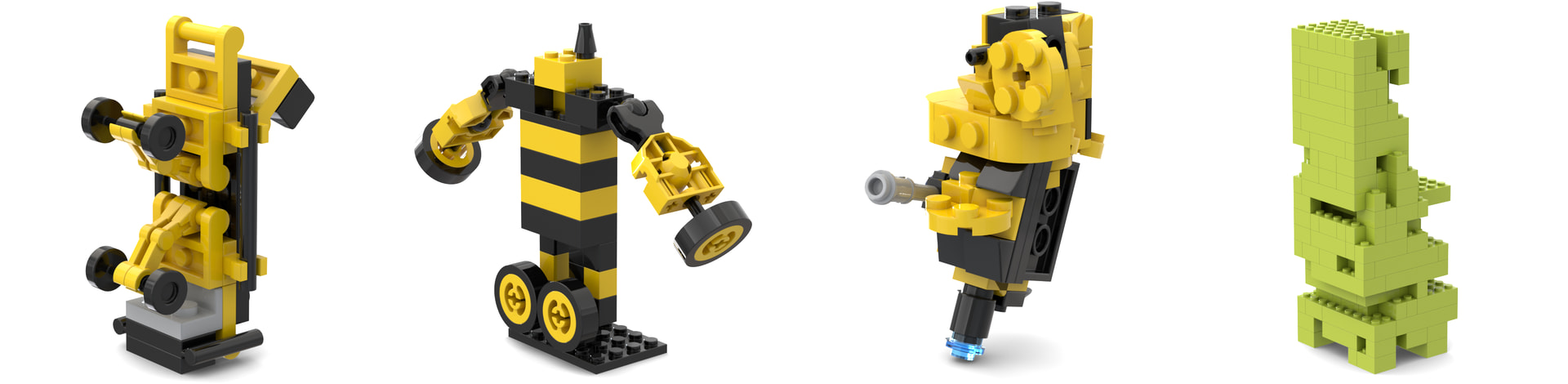}{This is a small, blocky LEGO robot model, primarily constructed from yellow and black bricks, featuring articulated limbs with wheel-like hands and feet, and presented from multiple angles to showcase its design.}
    \bnrow{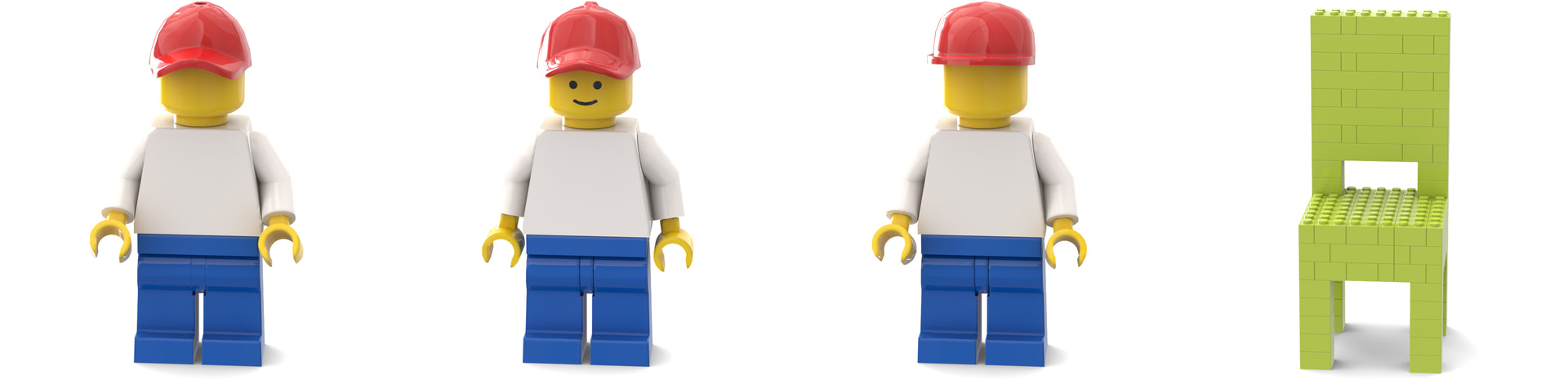}{This is a 3D model of a classic LEGO minifigure, featuring a yellow head and hands, a red baseball cap, a plain white torso, and blue legs, shown from multiple angles against a white background.}
    \vspace{2pt}\noindent%
    \bncol{GT}\bncol{GPT-5.6 Luna}\bncol{BrickNet-14B}\bncol{BrickGPT}
  \end{minipage}}%
  \begin{subfigure}[b]{\bnleftw}
    \usebox{\bnpanel}
    \subcaption{}\label{fig:bricknet:examples}
  \end{subfigure}\hfill
  \begin{subfigure}[b]{\dimexpr\textwidth-\bnleftw-8pt\relax}
    \parbox[b][\dimexpr\ht\bnpanel+\dp\bnpanel\relax][c]{\linewidth}{\centering\setlength{\tabcolsep}{4pt}%
    \begin{tabular}{@{}lccc@{}}
      \toprule
      Method & PE $\uparrow$ & SigLIP~2 $\uparrow$ & VQAScore $\uparrow$ \\
      \midrule
      BrickGPT & 0.157 & 0.052 & 0.050 \\
      BrickNet-0.6B & 0.279 & 0.603 & 0.557 \\
      BrickNet-1.7B & 0.282 & 0.631 & 0.593 \\
      BrickNet-4B & 0.283 & 0.639 & 0.615 \\
      BrickNet-8B & 0.284 & 0.647 & 0.608 \\
      BrickNet-14B & 0.283 & 0.625 & 0.602 \\
      \midrule
      GT & 0.315 & 0.826 & 0.748 \\
      \midrule
      GPT-5.6 Luna & \textbf{0.313} & \textbf{0.818} & \textbf{0.732} \\
      \bottomrule
    \end{tabular}}
    \subcaption{}\label{fig:bricknet:metrics}
  \end{subfigure}
  \caption{
\textbf{BrickNet validation.}
We evaluate GPT-5.6 Luna on the BrickNet text-to-assembly validation set~\citep{Kulits_2026_CVPR}.
While it was not specialized for this dataset, we find it exceeds task-specific baselines including BrickGPT~\citep{Pun_2025_ICCV} across metrics and nears the alignment of the ground-truth objects.
(a) Sample inputs and model outputs.
(b) Caption alignment following BrickNet's evaluation protocol.
PE~\citep{bolya2025pe} and SigLIP~2~\citep{tschannen2025siglip2} score the similarity between the embeddings of the renders and of the caption, and VQAScore~\citep{lin2024vqascore} the probability that a VQA model answers yes when asked whether the renders show the caption.
See \citet{Kulits_2026_CVPR} for details on metric computation.
}
\label{fig:bricknet}
\end{figure}

In text-conditioned assembly, a system receives a text prompt and is tasked with producing an assembly that depicts it.
The system must select part types from the library and assign each instance a color, position, and 3D rotation.

Prior work BrickNet trains a language model for this task on human-designed assemblies collected from online.
Its captions are generated by a vision--language model over renders of existing assemblies.
The validation set contains 512 such held-out captions from single objects of at most 100 parts each.

We evaluate GPT-5.6 Luna, a general coding agent, on the BrickNet validation prompts (\cref{fig:bricknet}).
The agent is given the environment described in \cref{ssec:brickagent}.
To match BrickNet, we limit the agent to 100 parts.
An output is scored by the alignment between its renders and the prompt, measured with PE~\citep{bolya2025pe}, SigLIP~2~\citep{tschannen2025siglip2}, and VQAScore~\citep{lin2024vqascore}.

Without having been trained on the dataset, the agent substantially outperforms task-specific baselines BrickNet and BrickGPT~\citep{Pun_2025_ICCV}, scoring within 0.02 of the held-out (GT) assemblies on each metric and effectively saturating the evaluation.
This motivates our design of a new benchmark.

\section{BrickBench}\label{sec:benchmark}
BrickBench is a benchmark of agentic text-conditioned LEGO-set design.
Given a text prompt, an agent must produce an assembly that satisfies semantic and design criteria and that can be physically built.
The benchmark consists of 300 prompts in three settings (\cref{ssec:settings}), an environment in which a coding agent can build, and an evaluation protocol for validity, semantic alignment, and design quality (\cref{ssec:metrics}).
Except for the ablation in \cref{ssec:ablation} and the two data-driven baselines, the agents work inside of the BrickAgent environment described in \cref{ssec:brickagent}.

\subsection{Settings}\label{ssec:settings}
The settings differ in the parts that an agent builds with.
In the \texttt{Model} setting, an agent may use at most 400 parts, while the \texttt{Set} setting requires 400--4000 parts be used.
Both draw from the full LDraw library as supported by BrickNet, 14,441 part types.
In the \texttt{Alt-Build} setting, an agent may use only the 783 pieces, of 104 part types, of the retail set 10698,\footnote{\url{https://www.bricklink.com/v2/catalog/catalogitem.page?S=10698}} containing mostly basic bricks and plates with a few windows, doors, wheels, and decorative elements.
An assembly that violates these limits fails the validity check.

The settings are intended to evaluate different abilities.
\texttt{Model} serves as the base evaluation and tests the creation of assemblies consisting primarily of a single component.
\texttt{Set}, whose size is chosen to mirror the requirements of the BrickLink Designer Program,\footnote{\url{https://www.bricklink.com/v3/designer-program/main.page}} tests whether an agent can build an assembly on the scale of a full-size retail set.
\texttt{Alt-Build} evaluates agent ability to design given a scarce, fixed part inventory.

\subsection{Prompts}\label{ssec:prompts}
Each evaluation setting has 100 prompts, with ten in each of ten topic categories.
They follow the BrickLink Designer Program, except that we split the Floral / Nature / Animal category into Floral / Nature and Animal.
The categories are Medieval / Castle, Vehicle / Boat / Airplane, Train, Space / Sci-Fi / Fantasy, Art / Object, Building, Floral / Nature, Animal, Pirate, and History / Period.
See \cref{fig:teaser,fig:grid-main} for prompt samples by setting.
The prompts were constructed through prompting of an LLM and were manually validated for conformity.
Prompts average 31 words in \texttt{Alt-Build}, 45 in \texttt{Model}, and 57 in \texttt{Set}.

We decompose each generation prompt into a question graph in the style of DSG~\citep{cho2024dsg}, which evaluates text-to-image generation by turning a prompt into atomic yes/no questions with an LLM and answering them with a VLM after organizing them by graph dependency.
The questions cover the entities named by the prompt, their attributes, and the relations among them.
With the graph we can avoid evaluating properties invalidated by other failures, such as checking the color of a knight's armor when the presence of a knight was not yet established.
See \cref{fig:question-types} for their distribution by type and \cref{fig:question-graph} for a full example graph, both in \Cref{app:prompts}.

\subsection{BrickAgent}\label{ssec:brickagent}
We introduce BrickAgent, an environment in which the agents build.
It contains the LDraw library, the BrickNet connector system, and tools for authoring, validating, and viewing an assembly programmatically.
The tools include functions to search the library for parts, to place parts by their positions or by connectors, to mirror and measure them, to define subassemblies and combine them, and to render an assembly from any viewpoint.
The provided validator includes the connectivity, collision, and stability tests of \cref{ssec:validity} and interpretably reports each failure with the parts involved.
This enables an agent to validate the physical requirements of an assembly prior to submission.

The prompt explains the environment with worked examples and states the setting's part requirements.
It informs the agent that experienced LEGO designers will judge the assembly against others by the standards of a published set.
In the ablation of \cref{ssec:ablation} the agent operates in the same container, with the harness CLI, LDraw library, and task requirements, but without the tools of BrickAgent.
We find that without these tools, frontier coding agents cannot reliably construct valid builds.

\subsection{Metrics}\label{ssec:metrics}
We score each system per setting, and additionally aggregate across all three (\texttt{Overall}).
An assembly is judged as valid if it meets the part requirements of its setting and is both collision-free and stable under the tests of \cref{ssec:validity}.
\texttt{Valid} is a proportion and is scored on every prompt.

\texttt{VQA} measures bounded semantic alignment with the prompt, following DSG.
We render the assembly from eight views using BrickNet's renderer, with consistent render settings.
We then ask Gemma~4 31B~\citep{gemmateam2026gemma4} each question of the graph in one call, with instructions to judge on what the views show and to answer yes or no.
The score of an assembly is the fraction of its questions satisfied, and \texttt{VQA} is the mean over assemblies.

\texttt{ELO} measures relative semantic alignment and design quality against other systems.
For each prompt, we form every pairing of assemblies from different systems and show the judge four views of each.
We ask two questions of each pair:
1) The alignment question gives the judge the prompt and asks which assembly matches it better.
2) The design question withholds the prompt and asks which assembly is better designed by the standard of an official set, judged on how the pieces are used rather than the subject.
Each pair is asked in both orderings to eliminate position bias.
\cref{fig:judge-prompts} in \Cref{app:judge-prompts} contains the judge prompts.
We fit a Bradley--Terry model~\citep{bradley1952rank} to the wins from each question and report them as ELO.
We anchor to a fixed set of references as noted in \cref{sec:evaluations}.
\texttt{ELO} is the average of \texttt{Align ELO} and \texttt{Design ELO} after rescaling to a common spread over the anchors.

\texttt{VQA} and \texttt{ELO} are scored on the assemblies generated by a system, without considering whether they are \texttt{Valid}.
\texttt{Cost} is the list price of the tokens consumed by a system for an assembly in US dollars.
For the two trained baselines it is the serving cost per assembly, at serverless token rates for the base model of BrickNet and as wall-clock time on a rented GPU instance for BrickGPT.

\section{Evaluations}\label{sec:evaluations}
\begin{table}[tp]
  \centering
  \caption{\textbf{BrickBench.}
We report metrics for each of the three evaluation settings and in the aggregate (\texttt{Overall}).
\texttt{Valid} requires that the assembly satisfies the part requirement of the setting and is both collision-free and stable.
\texttt{VQA} is the proportion of VQA questions satisfied (\cref{ssec:metrics}).
\texttt{ELO}, \texttt{Align ELO}, and \texttt{Design ELO} are ratings estimated from pairwise VLM judgments (\cref{ssec:metrics}).
\texttt{Cost} is the API cost in US dollars, or its equivalent for BrickGPT~\citep{Pun_2025_ICCV}, of producing one assembly.
*The \texttt{Valid} metric of BrickNet-14B~\citep{Kulits_2026_CVPR} doesn't include stability as the model doesn't produce assemblies in a canonical frame and the direction of gravity is undefined.
}
\label{tab:main}
  \footnotesize\setlength{\tabcolsep}{4pt}\renewcommand{\arraystretch}{0.93}\setlength{\aboverulesep}{0.15ex}\setlength{\belowrulesep}{0.25ex}
  \providecommand{\sysname}[2]{%
  \makebox[11pt][l]{\ifx\relax#1\relax\else\raisebox{-1.5pt}{\includegraphics[height=8pt]{fig/icons/#1}}\fi}\hspace{2pt}#2}
\begin{tabular*}{\textwidth}{@{\extracolsep{\fill}}lrrrrrr@{}}
  \toprule
  System & Valid $\uparrow$ & VQA $\uparrow$ & ELO $\uparrow$ & Align ELO $\uparrow$ & Design ELO $\uparrow$ & Cost (\$) $\downarrow$ \\
    \midrule
    \multicolumn{7}{@{}l}{\emph{Overall}} \\
    \midrule
    \sysname{openai}{GPT-6 Astra} & 1.00 & 0.954 & 1297 {\scriptsize$\pm$22} & 1295 {\scriptsize$\pm$27} & 1299 {\scriptsize$\pm$27} & 4.06 \\
    \sysname{openai}{GPT-6.1 Sol} & 1.00 & 0.945 & 1293 {\scriptsize$\pm$23} & 1301 {\scriptsize$\pm$27} & 1286 {\scriptsize$\pm$26} & 0.87 \\
    \sysname{anthropic}{Claude Opus 5.5} & 0.99 & 0.923 & 1249 {\scriptsize$\pm$23} & 1212 {\scriptsize$\pm$26} & 1287 {\scriptsize$\pm$27} & 6.33 \\
    \sysname{anthropic}{Claude Opus 5} & 1.00 & 0.908 & 1101 {\scriptsize$\pm$17} & 1095 {\scriptsize$\pm$20} & 1108 {\scriptsize$\pm$20} & 16.47 \\
    \sysname{qwen}{Qwen 3.8 Flash} & 0.90 & 0.852 & 1048 {\scriptsize$\pm$17} & 1040 {\scriptsize$\pm$21} & 1057 {\scriptsize$\pm$20} & 1.75 \\
    \sysname{openai}{GPT-5.6 Sol} & 1.00 & 0.859 & 1015 {\scriptsize$\pm$16} & 1041 {\scriptsize$\pm$19} & 988 {\scriptsize$\pm$19} & 1.02 \\
    \sysname{gemini}{Gemini 3.8 Flash} & 0.98 & 0.856 & 1014 {\scriptsize$\pm$17} & 1005 {\scriptsize$\pm$21} & 1024 {\scriptsize$\pm$20} & 3.70 \\
    \sysname{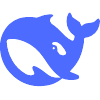}{DeepSeek V4.1 Flash} & 0.94 & 0.813 & 1006 {\scriptsize$\pm$17} & 998 {\scriptsize$\pm$20} & 1015 {\scriptsize$\pm$20} & 0.71 \\
    \sysname{openai}{GPT-5.6 Luna} & 1.00 & 0.792 & 898 {\scriptsize$\pm$16} & 917 {\scriptsize$\pm$19} & 879 {\scriptsize$\pm$20} & 0.75 \\
    \sysname{meta}{Muse Spark 1.3} & 0.80 & 0.718 & 868 {\scriptsize$\pm$20} & 859 {\scriptsize$\pm$24} & 878 {\scriptsize$\pm$23} & 6.68 \\
    \sysname{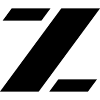}{GLM 5.3 Flash} & 0.92 & 0.590 & 752 {\scriptsize$\pm$23} & 750 {\scriptsize$\pm$25} & 753 {\scriptsize$\pm$29} & 0.54 \\
    \midrule
    \multicolumn{7}{@{}l}{\emph{Model}} \\
    \midrule
    \sysname{openai}{GPT-6 Astra} & 1.00 & 0.966 & 1266 {\scriptsize$\pm$38} & 1277 {\scriptsize$\pm$45} & 1254 {\scriptsize$\pm$45} & 4.08 \\
    \sysname{openai}{GPT-6.1 Sol} & 1.00 & 0.969 & 1269 {\scriptsize$\pm$36} & 1272 {\scriptsize$\pm$45} & 1265 {\scriptsize$\pm$42} & 0.92 \\
    \sysname{anthropic}{Claude Opus 5.5} & 1.00 & 0.952 & 1236 {\scriptsize$\pm$40} & 1216 {\scriptsize$\pm$46} & 1252 {\scriptsize$\pm$42} & 5.90 \\
    \sysname{anthropic}{Claude Opus 5} & 1.00 & 0.917 & 1083 {\scriptsize$\pm$30} & 1075 {\scriptsize$\pm$36} & 1089 {\scriptsize$\pm$31} & 16.72 \\
    \sysname{qwen}{Qwen 3.8 Flash} & 0.95 & 0.921 & 1012 {\scriptsize$\pm$27} & 1030 {\scriptsize$\pm$30} & 996 {\scriptsize$\pm$32} & 1.59 \\
    \sysname{openai}{GPT-5.6 Sol} & 1.00 & 0.908 & 1032 {\scriptsize$\pm$28} & 1051 {\scriptsize$\pm$34} & 1015 {\scriptsize$\pm$31} & 1.14 \\
    \sysname{gemini}{Gemini 3.8 Flash} & 0.98 & 0.894 & 1012 {\scriptsize$\pm$34} & 995 {\scriptsize$\pm$40} & 1028 {\scriptsize$\pm$37} & 4.05 \\
    \sysname{deepseek}{DeepSeek V4.1 Flash} & 0.95 & 0.857 & 1007 {\scriptsize$\pm$30} & 1013 {\scriptsize$\pm$33} & 1001 {\scriptsize$\pm$33} & 0.65 \\
    \sysname{openai}{GPT-5.6 Luna} & 1.00 & 0.821 & 892 {\scriptsize$\pm$30} & 897 {\scriptsize$\pm$34} & 889 {\scriptsize$\pm$37} & 0.99 \\
    \sysname{meta}{Muse Spark 1.3} & 0.86 & 0.811 & 848 {\scriptsize$\pm$31} & 845 {\scriptsize$\pm$38} & 851 {\scriptsize$\pm$36} & 5.99 \\
    \sysname{zai}{GLM 5.3 Flash} & 0.90 & 0.658 & 848 {\scriptsize$\pm$37} & 816 {\scriptsize$\pm$42} & 878 {\scriptsize$\pm$43} & 0.69 \\
    \sysname{}{BrickNet-14B} & 0.26 & 0.180 & 731 {\scriptsize$\pm$53} & 485 {\scriptsize$\pm$62} & 952 {\scriptsize$\pm$50} & 0.0006 \\
    \sysname{}{BrickGPT} & 1.00 & 0.024 & 338 {\scriptsize$\pm$45} & 129 {\scriptsize$\pm$64} & 528 {\scriptsize$\pm$60} & 0.04 \\
    \midrule
    \multicolumn{7}{@{}l}{\emph{Set}} \\
    \midrule
    \sysname{openai}{GPT-6 Astra} & 1.00 & 0.933 & 1305 {\scriptsize$\pm$35} & 1291 {\scriptsize$\pm$42} & 1316 {\scriptsize$\pm$46} & 5.15 \\
    \sysname{openai}{GPT-6.1 Sol} & 1.00 & 0.920 & 1295 {\scriptsize$\pm$41} & 1293 {\scriptsize$\pm$47} & 1292 {\scriptsize$\pm$45} & 1.05 \\
    \sysname{anthropic}{Claude Opus 5.5} & 0.99 & 0.892 & 1240 {\scriptsize$\pm$44} & 1174 {\scriptsize$\pm$45} & 1315 {\scriptsize$\pm$53} & 8.08 \\
    \sysname{anthropic}{Claude Opus 5} & 1.00 & 0.890 & 1100 {\scriptsize$\pm$27} & 1092 {\scriptsize$\pm$31} & 1109 {\scriptsize$\pm$34} & 18.80 \\
    \sysname{qwen}{Qwen 3.8 Flash} & 0.79 & 0.801 & 1090 {\scriptsize$\pm$38} & 1054 {\scriptsize$\pm$44} & 1131 {\scriptsize$\pm$41} & 2.26 \\
    \sysname{openai}{GPT-5.6 Sol} & 1.00 & 0.832 & 993 {\scriptsize$\pm$27} & 1019 {\scriptsize$\pm$32} & 963 {\scriptsize$\pm$32} & 1.16 \\
    \sysname{gemini}{Gemini 3.8 Flash} & 0.97 & 0.806 & 1012 {\scriptsize$\pm$31} & 1010 {\scriptsize$\pm$36} & 1013 {\scriptsize$\pm$36} & 4.26 \\
    \sysname{deepseek}{DeepSeek V4.1 Flash} & 0.87 & 0.786 & 1005 {\scriptsize$\pm$35} & 979 {\scriptsize$\pm$39} & 1036 {\scriptsize$\pm$41} & 0.87 \\
    \sysname{openai}{GPT-5.6 Luna} & 1.00 & 0.758 & 929 {\scriptsize$\pm$26} & 952 {\scriptsize$\pm$31} & 902 {\scriptsize$\pm$35} & 0.85 \\
    \sysname{meta}{Muse Spark 1.3} & 0.68 & 0.661 & 871 {\scriptsize$\pm$38} & 864 {\scriptsize$\pm$46} & 882 {\scriptsize$\pm$42} & 8.54 \\
    \sysname{zai}{GLM 5.3 Flash} & 0.89 & 0.523 & 695 {\scriptsize$\pm$33} & 738 {\scriptsize$\pm$41} & 649 {\scriptsize$\pm$45} & 0.61 \\
    \midrule
    \multicolumn{7}{@{}l}{\emph{Alt-Build}} \\
    \midrule
    \sysname{openai}{GPT-6 Astra} & 1.00 & 0.961 & 1325 {\scriptsize$\pm$37} & 1314 {\scriptsize$\pm$51} & 1336 {\scriptsize$\pm$44} & 2.95 \\
    \sysname{openai}{GPT-6.1 Sol} & 1.00 & 0.946 & 1326 {\scriptsize$\pm$35} & 1339 {\scriptsize$\pm$43} & 1312 {\scriptsize$\pm$46} & 0.64 \\
    \sysname{anthropic}{Claude Opus 5.5} & 0.99 & 0.925 & 1277 {\scriptsize$\pm$37} & 1248 {\scriptsize$\pm$41} & 1307 {\scriptsize$\pm$45} & 5.01 \\
    \sysname{anthropic}{Claude Opus 5} & 1.00 & 0.917 & 1126 {\scriptsize$\pm$31} & 1120 {\scriptsize$\pm$36} & 1132 {\scriptsize$\pm$36} & 13.90 \\
    \sysname{qwen}{Qwen 3.8 Flash} & 0.97 & 0.827 & 1051 {\scriptsize$\pm$27} & 1038 {\scriptsize$\pm$35} & 1064 {\scriptsize$\pm$36} & 1.39 \\
    \sysname{openai}{GPT-5.6 Sol} & 1.00 & 0.836 & 1021 {\scriptsize$\pm$31} & 1054 {\scriptsize$\pm$36} & 986 {\scriptsize$\pm$33} & 0.76 \\
    \sysname{gemini}{Gemini 3.8 Flash} & 1.00 & 0.867 & 1023 {\scriptsize$\pm$25} & 1010 {\scriptsize$\pm$31} & 1036 {\scriptsize$\pm$31} & 2.79 \\
    \sysname{deepseek}{DeepSeek V4.1 Flash} & 0.99 & 0.797 & 1007 {\scriptsize$\pm$29} & 1002 {\scriptsize$\pm$33} & 1014 {\scriptsize$\pm$34} & 0.60 \\
    \sysname{openai}{GPT-5.6 Luna} & 1.00 & 0.797 & 875 {\scriptsize$\pm$28} & 903 {\scriptsize$\pm$39} & 845 {\scriptsize$\pm$37} & 0.40 \\
    \sysname{meta}{Muse Spark 1.3} & 0.86 & 0.678 & 887 {\scriptsize$\pm$30} & 871 {\scriptsize$\pm$36} & 904 {\scriptsize$\pm$37} & 5.52 \\
    \sysname{zai}{GLM 5.3 Flash} & 0.97 & 0.591 & 686 {\scriptsize$\pm$36} & 688 {\scriptsize$\pm$47} & 684 {\scriptsize$\pm$47} & 0.31 \\
  \bottomrule
\end{tabular*}

\end{table}

\begin{figure}[tb]
  \centering
  \includegraphics[width=\linewidth]{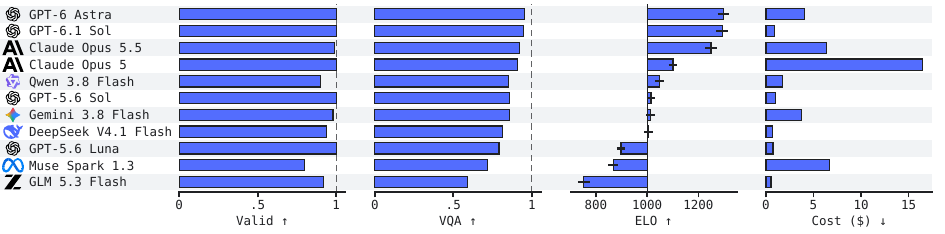}
  \caption{
  \textbf{BrickBench summary.}
  We report summary metrics averaged across the three evaluation settings.
  See \cref{tab:main} for disaggregated results.
  The ELO-plot lines represent 95\% intervals.
  }
  \label{fig:metric-bars}
\end{figure}

We evaluate eleven agents and two data-driven baselines.
The agents include GPT-6 Astra, GPT-6.1 Sol, GPT-5.6 Sol, GPT-5.6 Luna, Claude Opus 5.5, Claude Opus 5, Gemini 3.8 Flash, Qwen 3.8 Flash, DeepSeek V4.1 Flash, GLM 5.3 Flash, and Muse Spark 1.3.
Opus 5 and Opus 5.5 run in Claude Code, and the others in Codex CLI using the vendors' respective APIs.
Astra, both Opus models, and both Sol models are evaluated at \texttt{high} reasoning to bound cost while Luna runs at \texttt{max} and the others at \texttt{xhigh}.
Each agent is given a budget of 300 turns per prompt.
The baselines evaluated are BrickNet-14B~\citep{Kulits_2026_CVPR} and BrickGPT~\citep{Pun_2025_ICCV}, both restricted to the \texttt{Model} setting which is the only one applicable.
Nine of the agents form the reference set of the ELO scale (\cref{ssec:metrics}).
The baselines and ablated runs of \cref{ssec:ablation} are rated against them and do not affect the scale.
Opus 5.5 and GPT-6.1 Sol were released after the human studies of \cref{ssec:judge,ssec:human} were run, so they are scored on the automated metrics only.
Like the baselines, they are rated against the reference set, which leaves every other rating unchanged, and both fall within the range of ratings on which the judge is validated.

\subsection{Results}\label{ssec:results}
\providecommand{\gridcol}[3]{%
  \makebox[\dimexpr\textwidth/11\relax][c]{%
    \raisebox{-0.5\height}{\includegraphics[height=11.5pt]{fig/icons/#1}}\hspace{1.5pt}%
    \raisebox{-0.5\height}{\sffamily\fontsize{6}{7}\selectfont
      \begin{tabular}{@{}l@{}}#2\\#3\end{tabular}}}}
\providecommand{\gridlabels}{%
  \vspace{3pt}\noindent%
  \gridcol{openai}{GPT-6}{Astra}\gridcol{openai}{GPT-6.1}{Sol}\gridcol{anthropic}{Claude}{Opus 5.5}%
  \gridcol{anthropic}{Claude}{Opus 5}\gridcol{qwen}{Qwen 3.8}{Flash}\gridcol{openai}{GPT-5.6}{Sol}%
  \gridcol{gemini}{Gemini 3.8}{Flash}\gridcol{deepseek}{DeepSeek}{V4.1 Flash}\gridcol{openai}{GPT-5.6}{Luna}%
  \gridcol{meta}{Muse}{Spark 1.3}\gridcol{zai}{GLM 5.3}{Flash}}

\begin{figure}[p]
  \renewcommand{\tsrpromptfont}{\fontsize{6.5}{7.4}\selectfont}%
  \centering
  \teaserrow{Model}{tsrModel}{A green dumpster on four small black wheels has its lid pushed open by a tall cluster of red, orange, and yellow flames. A crumpled white sheet protrudes from one corner, and a tipped-over trash can lies beside it.}{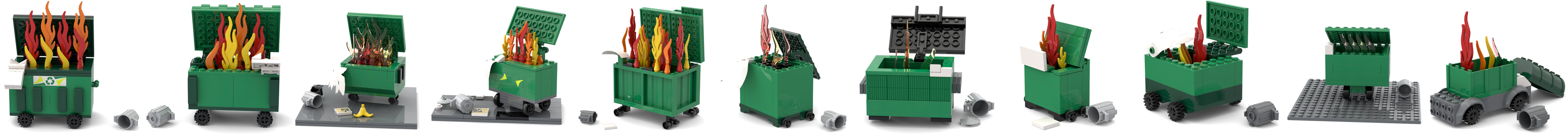}
  \teaserrow{Model}{tsrModel}{A cramped mailroom corner has a wall covered in white notices, envelopes, and photographs joined by tangled red strings. A wild-eyed, brown-haired man in a white shirt and loosened striped tie gestures toward the board. Cardboard boxes and overflowing stacks of mail crowd the floor beneath it.}{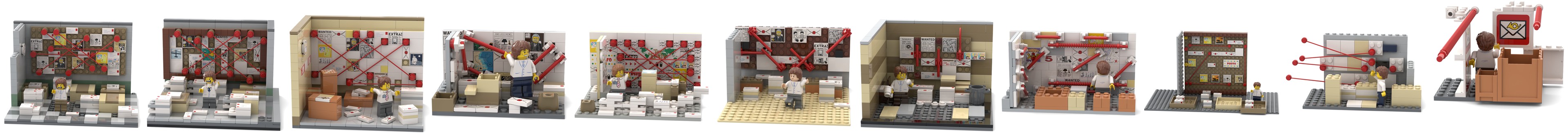}
  \teaserrow{Model}{tsrModel}{A blue dolphin leaps above a brilliant turquoise sea, its pale belly turned away from the rainbow. A bright rainbow arches behind its back between soft white clouds. The dolphin's tail curves back toward a little burst of white spray, while gentle blue waves fill the space beneath the rainbow.}{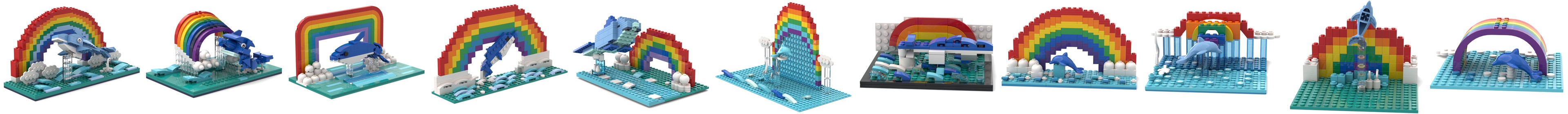}
  \teaserrow{Set}{tsrSet}{A masked swordsman dressed entirely in black sits opposite a small balding man at a rough stone table on a rocky hillside. Two silver goblets stand between them. Behind the balding man, a blindfolded blonde woman in a red dress sits with her hands bound. The swordsman rests one gloved hand near his goblet.}{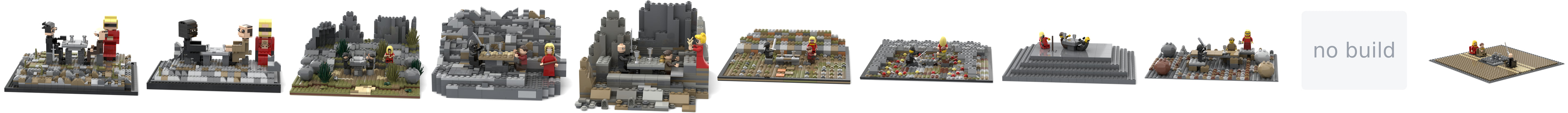}
  \teaserrow{Set}{tsrSet}{A gigantic blue toy locomotive with a round gray smiling face bursts through the side of a two-story suburban house. The locomotive is nearly as tall as the house, and its red buffer beam projects over the front lawn. Broken wall panels lie below it. Two people beside a small parked car look up at the engine.}{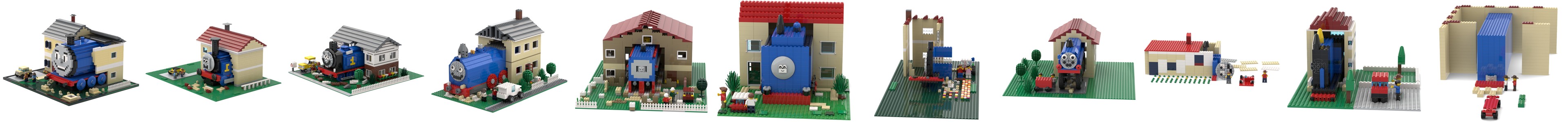}
  \teaserrow{Set}{tsrSet}{A blonde woman in a black off-the-shoulder top leans across a restaurant table, shouting and pointing directly at a white cat seated opposite her. A dark-haired woman stands behind her, gripping her shoulders and trying to pull her back. The cat sits upright on a dining chair with its ears splayed and its mouth slightly open, looking back at the pointing woman. A plate of green salad sits directly in front of the cat. Drinking glasses, cutlery, and folded napkins cover the table, with upholstered chairs and a low restaurant partition surrounding the confrontation.}{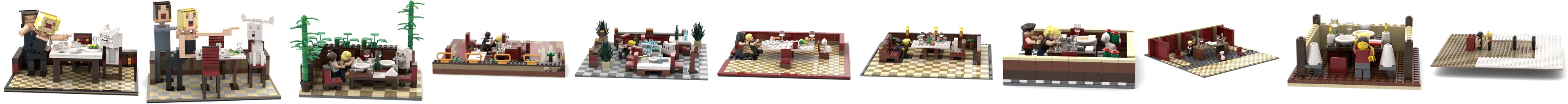}
  \teaserrow{Set}{tsrSet}{A huge, roughly spherical ball is covered in household objects pointing in every direction: a chair, a television, an umbrella, a frying pan, a banana, and several books. A tiny green figure with a long cylindrical head pushes against its base. Ahead of the ball, a traffic cone and a teapot remain on the floor.}{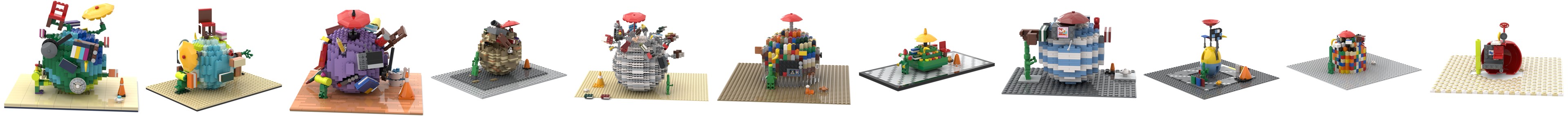}
  \teaserrow{Alt-Build}{tsrAlt}{A green sea serpent rises from the water beside a small pirate ship. Its neck bends over the bow, while its tail emerges from the water behind the stern.}{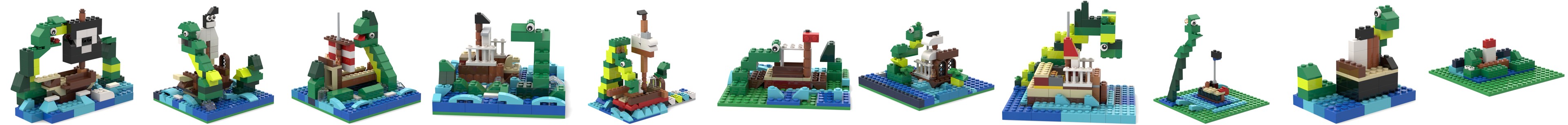}
  \teaserrow{Alt-Build}{tsrAlt}{A tiny blue-armored knight braces behind a shield and points a lance at a snail twice his height. The snail has a large spiral shell and two raised eye stalks. They face each other across a grassy patch.}{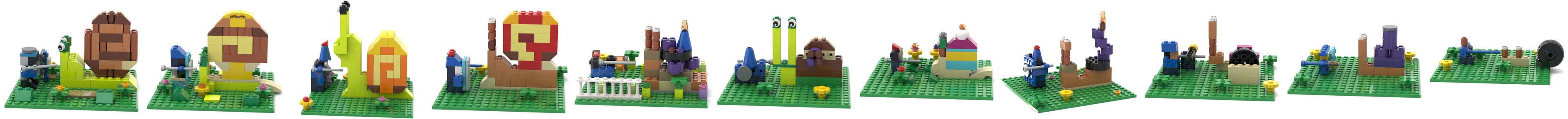}
  \teaserrow{Alt-Build}{tsrAlt}{A steam locomotive crossing a bridge over a river. The bridge has a single arch directly beneath the engine.}{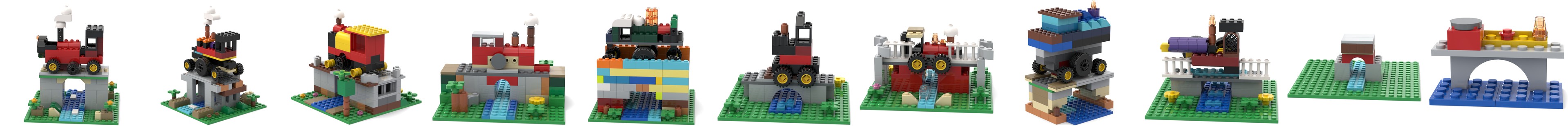}
  \gridlabels
  \caption{
  \textbf{BrickBench samples.}
  We show one assembly from each of the eleven agents for three prompts in each setting.
  We observe the agents vary in their interpretation of the prompts.
  One assembly is missing for Muse because it did not deliver one within the turn budget.
}
\label{fig:grid-main}
\end{figure}

\begin{figure}[tb]
  \centering
  \captionsetup[subfigure]{skip=2pt}%
  \begin{subfigure}[b]{0.49\textwidth}
    \centering
    \includegraphics[width=\linewidth]{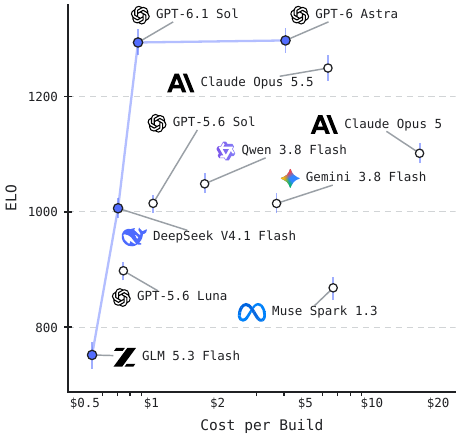}
    \subcaption{\texttt{ELO} vs.~\texttt{Cost}}\label{fig:elo-cost}
  \end{subfigure}\hfill
  \begin{subfigure}[b]{0.49\textwidth}
    \centering
    \includegraphics[width=\linewidth]{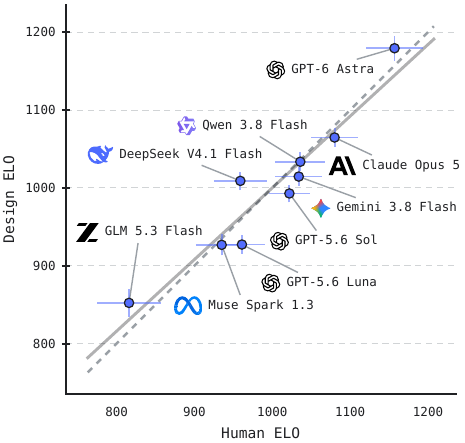}
    \subcaption{Validation of \texttt{Design ELO}}\label{fig:elo-human}
  \end{subfigure}
  \caption{
  \textbf{ELO.}
  (a) We plot \texttt{ELO} against mean assembly cost.
  (b) To validate our model-based \texttt{Design ELO} computation, we perform a perceptual study where participants are tasked with judging design quality between a pair of assemblies for a given prompt.
  The prompt isn't given to the judge, and they make their decision based only on the images.
  We find that the rankings agree on 34/36 agent pairs (Kendall $\tau=0.89$), suggesting the metric is an effective proxy for human judgment.
  The study covers the nine reference agents; Opus 5.5 and GPT-6.1 Sol were evaluated after it.
  }
  \label{fig:elo-pair}
\end{figure}

We report each metric, per setting and overall, in \cref{tab:main}, and plot the overall columns in \cref{fig:metric-bars}.
We observe that validity is nearly solved when provided the environment.
Five agents deliver a valid assembly for each prompt and none fall below a \texttt{Valid} proportion of 0.80 overall.
Of the 137 invalid assemblies across the reference set, 44 are prompts with no delivered assembly.
The others fail on stability (41), collisions (31), and the part requirement (21).

\texttt{VQA} separates the reference set gently, from 0.59 for GLM to 0.95 for Astra.
\texttt{Design ELO} does so more sharply, with 545 ELO points between GLM and Astra, a gap that can be interpreted as a judge preferring Astra's assembly in around 96\% of pairs.
Within the reference set, Astra leads on every metric, and Opus 5 is ranked second while incurring four times the cost.
Sol, Gemini, Qwen, and DeepSeek form a middle within 41 ELO points of one another.
Of the two later agents, GPT-6.1 Sol matches Astra on \texttt{ELO} (1293 against 1297, intervals overlapping) at a fifth of its cost, and Opus 5.5 rates 148 points above Opus 5 at 40\% of its cost, inside Astra's interval on \texttt{Design ELO} but 83 points lower on \texttt{Align ELO}.
Alignment varies more with the kind of question than with the subject (\cref{fig:vqa-radars} in \Cref{app:results}).
No prompt category is consistently harder than the others, and the two top agents score at least 0.90 in every one.
Every agent satisfies at least 91\% of the questions about color, but texture is satisfied 49--83\% of the time, and object type and state are the next hardest.
The stronger agents also build larger and more varied assemblies (\cref{tab:stats} in \Cref{app:results}).
In \texttt{Model}, Astra averages 217 parts of 35 types in 9 colors, and GLM 27 parts of 10 types in 5 colors.
In \texttt{Set}, Qwen, DeepSeek, Muse, and GLM average 23 to 40 connected components per assembly, against 2.2 to 4.2 for Astra and both Sol and Opus models, so their scenes consist of far more disconnected pieces.
\cref{fig:metric-bars-t1,fig:metric-bars-t3} in \Cref{app:results} plot \cref{tab:main,tab:stats} by setting.

The reference set spans a factor of 30 in cost, and rank does not follow price.
Muse, for instance, costs nine times what DeepSeek costs and rates 138 points lower.
See \cref{fig:elo-cost} for a plot of \texttt{ELO} against \texttt{Cost}.
While \texttt{Align ELO} and \texttt{Design ELO} agree for most agents, Sol scores 53 points higher on alignment than design.
Both trained baselines fall drastically below every agent.
BrickNet-14B is collision-free on only 26\% of prompts, while BrickGPT, though valid on each prompt, satisfies 2\% of the questions.
See \cref{fig:grid-main} for sample visuals, and \cref{fig:grid-appendix} in \Cref{app:results} for ten more prompts.
We observe that the settings differ in what breaks.
In terms of validity, \texttt{Set} is the hardest, with validity dropping to 0.68 for Muse, 0.79 for Qwen, and 0.87 for DeepSeek.
Alignment falls for every agent, and part counts crowd the 400-part floor (\cref{fig:parts-density} in \Cref{app:results}), so many agents build the smallest scene allowed by the setting.
\texttt{Alt-Build} costs the least and maintains the highest validity.
Alignment is highest in \texttt{Model}, the only setting in which the baselines apply.

\subsection{Environment Ablation}\label{ssec:ablation}
\begin{table}[tb]
  \centering
  \caption{
    \textbf{Environment ablation.}
    We ablate the effect of the BrickAgent environment, pooled over the three evaluation settings (\texttt{Overall}; 300 prompts).
    While both agents consistently produce valid assemblies when given tools, they struggle without them to varying degrees: GPT-6 Astra is somewhat robust but its assemblies drop to being valid only 40\% of the time while GPT-5.6 Luna's fall to less than a percent.
    The environment improves the VQA of Astra but lowers its \texttt{ELO} within noise.
    Without it, Luna improves on both, suggesting the buildability constraint limits expressivity.
}
\label{tab:ablation}
  \small
  \providecommand{\sysname}[2]{%
  \makebox[11pt][l]{\ifx\relax#1\relax\else\raisebox{-1.5pt}{\includegraphics[height=8pt]{fig/icons/#1}}\fi}\hspace{2pt}#2}
\begin{tabular}{@{}lrrrrrr@{}}
  \toprule
  System & Valid $\uparrow$ & Collisions $\downarrow$ & Conn.\ Components & VQA $\uparrow$ & ELO $\uparrow$ & Cost (\$) \\
  \midrule
    \sysname{openai}{GPT-6 Astra} & 1.00 & 0.00 & 2.21 & 0.954 & 1297 {\scriptsize$\pm$22} & 4.06 \\
    \quad without BrickAgent & 0.40 & 7.19 & 6.12 & 0.940 & 1309 {\scriptsize$\pm$24} & 4.66 \\
    \sysname{openai}{GPT-5.6 Luna} & 1.00 & 0.00 & 4.60 & 0.792 & 898 {\scriptsize$\pm$16} & 0.75 \\
    \quad without BrickAgent & $<$0.01 & 155.51 & 83.00 & 0.813 & 960 {\scriptsize$\pm$20} & 0.60 \\
  \bottomrule
\end{tabular}

\end{table}

We ablate the BrickAgent environment in \cref{tab:ablation}, with the same agent CLI (Codex CLI), container, and task requirements, but with only the LDraw library and a primer on the LDraw file format in place of its tools.
Astra remains valid on 40\% of prompts while Luna produces only one valid build in 300.
Their assemblies average 7 and 156 colliding pairs, respectively, and fall apart into 6 and 83 connected components against 2 and 5 with the environment.

Alignment and design do not fail, however.
Astra's \texttt{VQA} drops by 0.01 and its \texttt{ELO} moves within the interval.
Both of Luna's metrics rise, by 0.02 in \texttt{VQA} and by 62 ELO points.
Without the restriction of buildability, Luna makes assemblies that depict the prompt as well and look better, but that cannot be built, demonstrating the importance of relevant verifiers.
Doing so costs Astra nothing on alignment or design, but costs Luna a little of both.

\subsection{Validation of the Design Judge}\label{ssec:judge}
To confirm the validity of \texttt{Design ELO}, which rests on a VLM judge, we evaluate against human ratings.
We recruit participants on Prolific and show each 40 pairs of same-prompt assemblies, with four views per assembly.
We collect 4,080 judgments from 102 raters.
We fit the human judgments with the same Bradley--Terry model as \texttt{ELO} (\cref{ssec:metrics}), and compare the ratings against \texttt{Design ELO} in \cref{fig:elo-human}.
The rankings agree on 34 of the 36 pairs among the agents, with a Kendall $\tau$ of 0.89.
This follows findings from prior work which demonstrated alignment between VLM and human preferences on text-to-3D generation~\citep{wu2024gpt4vision}.
The judge is more decisive than the rater pool, so we align the two by a constant factor.
Opus 5.5 and GPT-6.1 Sol were released after this study and are not part of it.

\subsection{Comparison with Human Designs}\label{ssec:human}
\newsavebox{\forumplot}
\newsavebox{\forummosaic}
\newlength{\forumgap}\setlength{\forumgap}{24pt}
\begin{figure}[tb]
  \centering
  \captionsetup[subfigure]{skip=2pt}%
  \sbox{\forumplot}{\includegraphics[height=\dimexpr(\textwidth-\forumgap)*1000/3423\relax]{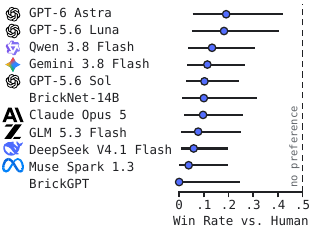}}%
  \sbox{\forummosaic}{\includegraphics[height=\dimexpr(\textwidth-\forumgap)*1000/3423\relax]{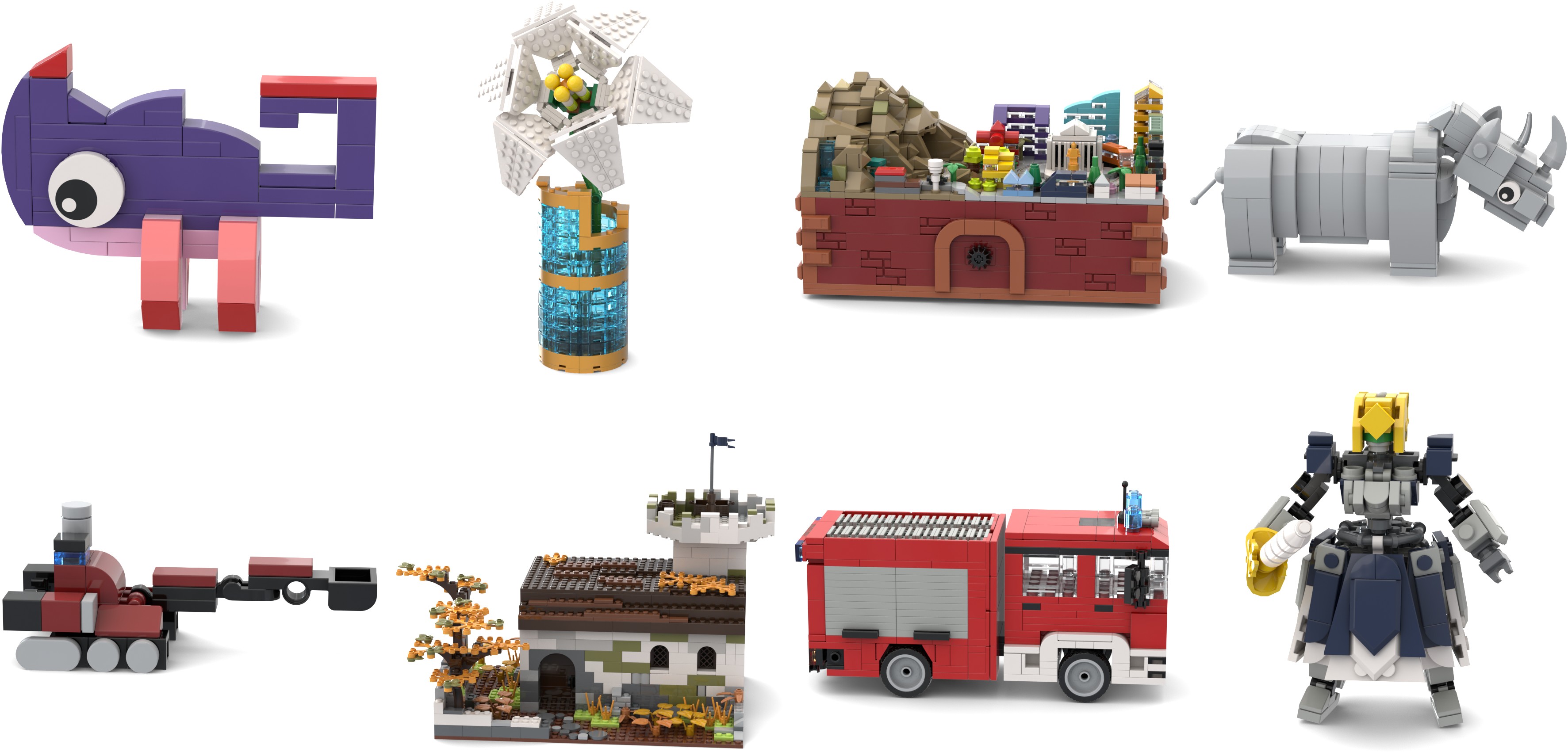}}%
  \begin{subfigure}[b]{\wd\forumplot}
    \usebox{\forumplot}
    \subcaption{}\label{fig:forum:rates}
  \end{subfigure}\hfill
  \begin{subfigure}[b]{\wd\forummosaic}
    \usebox{\forummosaic}
    \subcaption{}\label{fig:forum:models}
  \end{subfigure}
  \vspace{-0.3cm}
  \caption{
  \textbf{Human-design evaluation.}
We perform a perceptual study comparing agent designs with human designs.
For each assembly in the \texttt{Model} and \texttt{Set} settings, we select a random assembly from the BrickNet dataset with a comparable number of parts and ask a human rater to identify the one made by a person.
We find that raters are able to distinguish between the agent-designed and human-designed assemblies (a); the study covers the nine reference agents and the two baselines.
While agent-designed assemblies often satisfy the prompt criteria, there is a clear gap in fine-grained design to those produced by humans (b).
}
\vspace{-0.3cm}
\label{fig:forum}
\end{figure}

We asked people familiar with LEGO design to tell agent builds from human ones (\cref{fig:forum}).
For assemblies in the \texttt{Model} and \texttt{Set} settings, we drew part-count-matched human-designed models from the BrickNet dataset, and asked which of the two a person made.
The drawn samples are not prompt-matched.
No human designs exist for the BrickBench prompts, so the pairs are matched on part count alone, and \cref{fig:forum:models} shows human models of other subjects.
A prompt-matched comparison would require commissioning designs, which we leave to future work.
As in \cref{ssec:judge}, the study predates Opus 5.5 and GPT-6.1 Sol.
Five raters provided 360 ratings altogether, and selected the human model in 323 of them.
No agent was taken for a human designer more than one time in five (\cref{fig:forum:rates}), and every agent's rate falls significantly below chance.
GPT-6 Astra, the agent most often taken for a person, was chosen in 4 of its 21 pairs.
We show reference examples of human-designed models in \cref{fig:forum:models}.
While agent assemblies satisfy the prompt well, they lack the part usage, proportion, and surface detail that mark human design.

\section{Discussion and Limitations}\label{sec:discussion}
We report four findings from our evaluations.
First, we find that data-driven models built for the task are subsumed by general agents.
A general coding agent exceeds models on their own benchmark (\cref{ssec:text-assembly}) and on ours.
Cost, however, remains an axis for future work.
BrickNet-14B produces an assembly for six hundredths of a cent, while the Astra does so in \$4.06.
Second, prompt adherence saturates.
The best-performing agent satisfies 95\% of the VQA questions asked of it, and absolute alignment does not strongly separate the top performers.
Headroom lies instead in design quality.
Third, agent performance is dependent on the environment.
The same model produces consistently buildable assemblies with BrickAgent, and unbuildable ones without it (\cref{tab:ablation}).
Fourth, a gap remains to humans.
Participants familiar with LEGO design filtered agent-designed assemblies from human ones nine out of ten times (\cref{fig:forum}).
We interpret this as a measure of design quality: agents satisfy requirements but fall short of the standards of a designed set.

Our simulator of physical validity is a proxy and not a complete measure of structural stability.
Connections are not tested for strength, so an assembly that balances passes even when a physical one might sag or come apart (\cref{ssec:validity}).
Some works exist that attempt to build force models over stud connections~\cite{wassmann2012maximum} to evaluate FEM~\citep{pletz2023brickfem}.
None, however, is general, and integrating such would severely limit the available part catalog.
Until a general one exists, our bar for physical evaluation is necessary but not sufficient.

\section{Conclusion}\label{sec:conclusion}
We introduced BrickBench, a benchmark for agentic text-conditioned LEGO-set design.
The domain gives the task a floor that a simulator can verify and an open ceiling on design quality.
It includes 300 prompts in three settings with different constraints, a reference environment in which agents build, BrickAgent, and a set of metrics that score build validity, alignment, and design confirmed by human raters.
Nine frontier agents build valid assemblies for most prompts and satisfy the majority of what each requests, displacing task-specific data-driven models at an increase of 900--6,800 times the cost.
While doing so, we highlight the open challenge of narrowing the gap to human design.

\clearpage
{
\small
\bibliographystyle{iclr2027_conference}
\bibliography{main}
}

\clearpage
\appendix
\setcounter{figure}{0}\setcounter{table}{0}
\renewcommand{\thefigure}{A\arabic{figure}}\renewcommand{\thetable}{A\arabic{table}}
\renewcommand{\theHfigure}{A\arabic{figure}}\renewcommand{\theHtable}{A\arabic{table}}
{%
    \centering
    {\titlefont\textls[10]{\textbf{\wordmark}: Evaluating Agentic Brick Design}\par}%
    \vspace{5mm}
    {\Large Appendix} \\[10mm]
}
\section{Prompts and Question Graphs}\label{app:prompts}

\begin{figure}[htb]
  \centering
  \includegraphics[width=\linewidth]{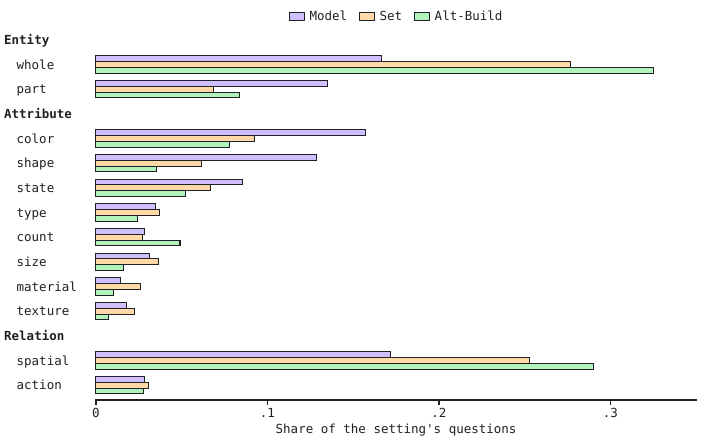}
  \caption{
  \textbf{VQA question types.}
  We plot the share of each setting's questions by DSG type.
  }
  \label{fig:question-types}
\end{figure}

\cref{fig:question-types} shows the share of each setting's questions by kind and subtype.
\texttt{Set} graphs are the largest, at 24 questions per prompt on average, because their prompts often describe scenes with multiple objects.

\begin{figure}[htb]
  \centering
  \noindent\tikz\node[tsrbox, line width=0.6pt, fill=tsrSet, text width=\dimexpr\textwidth-5pt\relax, inner xsep=2pt,
             inner ysep=1.5pt, align=left, font=\sffamily\fontsize{7}{8}\selectfont]
    {{\tsrmono\fontsize{7}{8}\selectfont Set}\enspace A brown dog with floppy black ears and a small brown hat sits at a table in a burning room. A white coffee mug rests in front of it. Flames climb the curtains and surround the chair, but the dog sits upright with its paws resting calmly on the tabletop. The front of the room is open to reveal the scene.};%
  \par\vspace{3pt}\nointerlineskip\noindent
  \includegraphics[width=\linewidth]{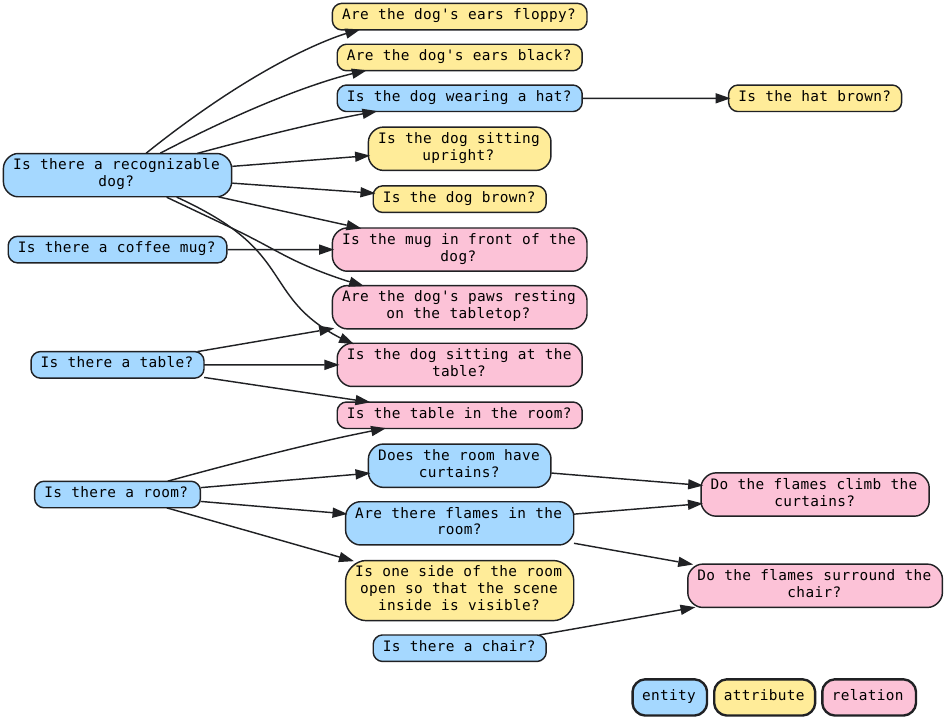}
  \caption{
  \textbf{Question-graph sample.}
We plot a sample question graph for a set prompt.
An arrow from one question to another indicates the second is satisfied only if the judge answers yes to the first.
  }
  \label{fig:question-graph}
\end{figure}

\cref{fig:question-graph} shows one \texttt{Set} graph in full, with the prompt it was generated from.
Its roots are the entities the prompt names, and every attribute and relation question hangs from the questions that establish what it refers to, so a missing entity fails every subsequent question about it.

\section{Judge Prompts}\label{app:judge-prompts}

\cref{fig:judge-prompts} reproduces the three prompts of \cref{ssec:metrics} as the judge receives them.
The two pairwise prompts differ in what the judge is told.
The alignment prompt contains the prompt and asks which assembly covers more of it, whatever its build quality.
The design prompt withholds the prompt and asks which assembly is better designed, whatever it depicts.
The VQA prompt lists every question of the graph, tells the judge to answer from the views alone, and requires a yes or a no on each.
\providecolor{tsrGray}{HTML}{F1F3F5}
\providecommand{\promptbox}[2]{%
  \noindent\tikz\node[tsrbox, line width=0.6pt, fill=tsrGray, text width=\dimexpr\textwidth-6.6pt\relax, inner xsep=3pt,
             inner ysep=2.5pt, align=left, font=\sffamily\fontsize{8}{9.6}\selectfont]
    {{\tsrmono\fontsize{8}{9.6}\selectfont #1}\enspace #2};%
  \par\vspace{3pt}\nointerlineskip}

\begin{figure}[htb]
  \centering
  \promptbox{Design}{The first image shows every view of one build, tiled. The second image shows the other build the same way.\\[3pt]
  You are shown two LEGO builds. The first 4 images are Build A, photographed from 4 angles. The next 4 images are Build B, the same 4 angles.\\[3pt]
  Pick the build that looks better designed, judged against the build quality you would expect from an official LEGO set. Look at how well the pieces are used, not at the subject: what a build depicts carries no weight here, and neither does how appealing you find it.\\[3pt]
  Both builds were rendered by the same software with identical lighting and background, so image quality carries no information.\\[3pt]
  Answer with exactly one character, A or B. No other text.}
  \promptbox{Alignment}{The first image shows every view of one build, tiled. The second image shows the other build the same way.\\[3pt]
  Two people were given the same description and each built it out of LEGO.\\[3pt]
  Description:\\
  {[}the prompt{]}\\[3pt]
  The first 4 images are Build A, photographed from 4 angles. The next 4 images are Build B, the same 4 angles.\\[3pt]
  Pick the build that better matches the description. Work through the description point by point: the subject, the parts it says the build should have, their arrangement, and any colours, counts or relations it states. A build that covers more of what was asked for wins, whatever its build quality.\\[3pt]
  Answer with exactly one character, A or B. No other text.}
  \promptbox{VQA}{You are shown 8 views of a single LEGO build, photographed from 8 angles around it.\\[3pt]
  It was built from this brief:\\
  {[}the prompt{]}\\[3pt]
  Answer the numbered questions below about THIS build. Use the views together -- a feature need not be visible in every view. Judge only what you can see: do not assume a detail is present because the brief asks for it, and do not infer hidden details from how such an object is usually made. Keep one consistent assignment of objects to roles across all the questions.\\[3pt]
  Say yes if the views support it and no if they do not. You must choose one of the two for every question; there is no third option. Where the evidence is thin, decide on the balance of what the eight views show.\\[3pt]
  {[}the questions, numbered{]}\\[3pt]
  End your reply with exactly {[}the number of questions{]} lines, nothing after them, one per question:\\[3pt]
  \textless number\textgreater: \textless yes\textbar no\textgreater}
  \caption{
  \textbf{Judge prompts.}
  The three prompts of \cref{ssec:metrics} as the judge receives them.
  For the two pairwise questions, the four views of each assembly are tiled into one image and the two images precede the text, and the design question never contains the prompt.
  For \texttt{VQA}, the eight views precede the text as separate images, and the bracketed fields are filled in for each assembly.
  }
  \label{fig:judge-prompts}
\end{figure}

\section{Additional Results}\label{app:results}

\cref{tab:stats} reports build statistics per setting.
Astra uses the most parts in every setting and Qwen the second most.
In \texttt{Set}, Qwen, DeepSeek, Muse, and GLM build scenes of 20 to 40 disconnected components on average, while Astra, Opus, Gemini, and Sol keep theirs under five.
\begin{table}[htbp]
  \centering
  \caption{
  \textbf{Further assembly statistics.}
  We report further statistics of the assemblies per setting.
  \texttt{Parts} is the average number of parts in an assembly.
  \texttt{Unique} is the (average) number of unique parts.
  \texttt{Colors} is the number of colors.
  \texttt{Collisions} is the number of parts that nonphysically collide (overlap) with one another.
  \texttt{Stable} is the proportion of assemblies that remain static under simulation, with every part moving under 3 LDU.
  \texttt{Conn.\ Components} is the number of graph components after BrickNet parsing; a valid assembly may have multiple components in order to represent multiple objects.
}
\label{tab:stats}
  \small\setlength{\tabcolsep}{4pt}\renewcommand{\arraystretch}{1.0}\setlength{\aboverulesep}{0.15ex}\setlength{\belowrulesep}{0.25ex}
  \providecommand{\sysname}[2]{%
  \makebox[11pt][l]{\ifx\relax#1\relax\else\raisebox{-1.5pt}{\includegraphics[height=8pt]{fig/icons/#1}}\fi}\hspace{2pt}#2}
\begin{tabular*}{\textwidth}{@{\extracolsep{\fill}}lrrrrrr@{}}
  \toprule
  System & Parts & Unique & Colors & Collisions $\downarrow$ & Stable $\uparrow$ & Conn.\ Components \\
    \midrule
    \multicolumn{7}{@{}l}{\emph{Model}} \\
    \midrule
    \sysname{openai}{GPT-6 Astra} & 217 & 35.4 & 9.4 & 0.00 & 1.000 & 1.6 \\
    \sysname{openai}{GPT-6.1 Sol} & 230 & 36.4 & 9.1 & 0.00 & 1.000 & 1.2 \\
    \sysname{anthropic}{Claude Opus 5.5} & 182 & 38.2 & 9.1 & 0.00 & 1.000 & 1.3 \\
    \sysname{anthropic}{Claude Opus 5} & 110 & 27.7 & 7.7 & 0.00 & 1.000 & 1.3 \\
    \sysname{qwen}{Qwen 3.8 Flash} & 125 & 23.8 & 8.3 & 0.10 & 0.979 & 4.8 \\
    \sysname{openai}{GPT-5.6 Sol} & 83 & 20.4 & 8.0 & 0.00 & 1.000 & 1.2 \\
    \sysname{gemini}{Gemini 3.8 Flash} & 95 & 25.6 & 9.6 & 0.00 & 0.990 & 1.2 \\
    \sysname{deepseek}{DeepSeek V4.1 Flash} & 88 & 19.3 & 7.0 & 0.56 & 0.969 & 3.1 \\
    \sysname{openai}{GPT-5.6 Luna} & 41 & 14.8 & 6.8 & 0.00 & 1.000 & 2.0 \\
    \sysname{meta}{Muse Spark 1.3} & 56 & 14.5 & 6.7 & 0.46 & 0.925 & 2.0 \\
    \sysname{zai}{GLM 5.3 Flash} & 27 & 10.3 & 4.9 & 0.86 & 0.928 & 1.7 \\
    \sysname{}{BrickNet-14B} & 40 & 14.5 & 4.8 & 24.80 & -- & 1.0 \\
    \sysname{}{BrickGPT} & 139 & 7.1 & 1.0 & 0.00 & 1.000 & 2.4 \\
    \midrule
    \multicolumn{7}{@{}l}{\emph{Set}} \\
    \midrule
    \sysname{openai}{GPT-6 Astra} & 1511 & 46.8 & 17.0 & 0.00 & 1.000 & 2.2 \\
    \sysname{openai}{GPT-6.1 Sol} & 1625 & 49.1 & 16.7 & 0.00 & 1.000 & 3.5 \\
    \sysname{anthropic}{Claude Opus 5.5} & 1112 & 68.5 & 17.5 & 0.00 & 1.000 & 3.8 \\
    \sysname{anthropic}{Claude Opus 5} & 865 & 48.6 & 14.6 & 0.00 & 1.000 & 4.2 \\
    \sysname{qwen}{Qwen 3.8 Flash} & 1118 & 38.8 & 17.4 & 0.02 & 0.929 & 32.8 \\
    \sysname{openai}{GPT-5.6 Sol} & 698 & 29.9 & 14.8 & 0.00 & 1.000 & 3.3 \\
    \sysname{gemini}{Gemini 3.8 Flash} & 581 & 36.5 & 17.3 & 0.02 & 0.980 & 1.7 \\
    \sysname{deepseek}{DeepSeek V4.1 Flash} & 726 & 33.5 & 12.9 & 6.94 & 0.888 & 36.7 \\
    \sysname{openai}{GPT-5.6 Luna} & 511 & 26.2 & 13.2 & 0.00 & 1.000 & 10.2 \\
    \sysname{meta}{Muse Spark 1.3} & 500 & 19.2 & 11.0 & 9.96 & 0.800 & 22.6 \\
    \sysname{zai}{GLM 5.3 Flash} & 568 & 16.5 & 9.0 & 0.80 & 0.908 & 40.2 \\
    \midrule
    \multicolumn{7}{@{}l}{\emph{Alt-Build}} \\
    \midrule
    \sysname{openai}{GPT-6 Astra} & 110 & 30.0 & 15.4 & 0.00 & 1.000 & 2.9 \\
    \sysname{openai}{GPT-6.1 Sol} & 110 & 30.8 & 15.4 & 0.00 & 1.000 & 2.0 \\
    \sysname{anthropic}{Claude Opus 5.5} & 103 & 27.5 & 14.2 & 0.00 & 1.000 & 3.3 \\
    \sysname{anthropic}{Claude Opus 5} & 86 & 25.1 & 12.9 & 0.00 & 1.000 & 1.8 \\
    \sysname{qwen}{Qwen 3.8 Flash} & 105 & 23.3 & 16.0 & 0.00 & 1.000 & 5.1 \\
    \sysname{openai}{GPT-5.6 Sol} & 52 & 19.3 & 12.1 & 0.00 & 1.000 & 1.4 \\
    \sysname{gemini}{Gemini 3.8 Flash} & 64 & 24.7 & 14.2 & 0.00 & 1.000 & 1.4 \\
    \sysname{deepseek}{DeepSeek V4.1 Flash} & 73 & 18.1 & 12.2 & 0.00 & 1.000 & 4.7 \\
    \sysname{openai}{GPT-5.6 Luna} & 40 & 17.9 & 11.2 & 0.00 & 1.000 & 1.6 \\
    \sysname{meta}{Muse Spark 1.3} & 49 & 14.8 & 11.0 & 0.63 & 0.945 & 2.0 \\
    \sysname{zai}{GLM 5.3 Flash} & 25 & 11.1 & 9.4 & 0.01 & 0.980 & 1.5 \\
  \bottomrule
\end{tabular*}

\end{table}

\begin{figure}[htb]
  \centering
  \captionsetup[subfigure]{skip=2pt}
  \begin{subfigure}{\textwidth}
    \centering
    \includegraphics[width=\linewidth]{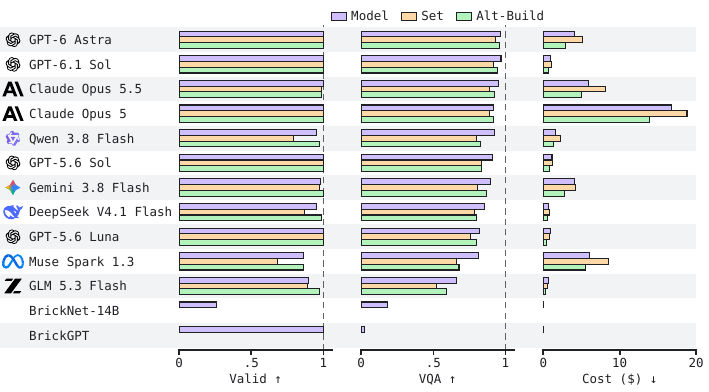}
    \subcaption{}\label{fig:metric-bars-t1:a}
  \end{subfigure}\\[4pt]
  \begin{subfigure}{\textwidth}
    \centering
    \includegraphics[width=\linewidth]{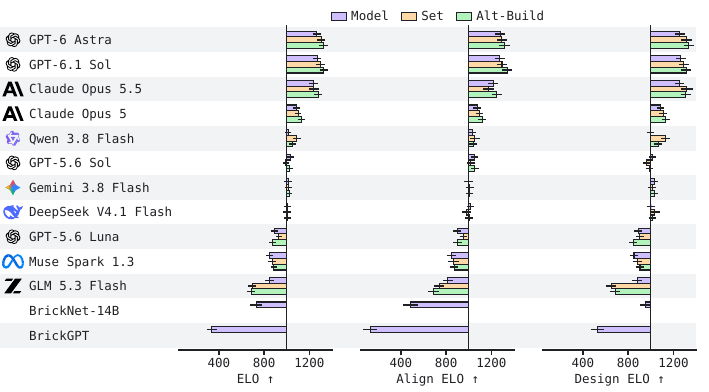}
    \subcaption{}\label{fig:metric-bars-t1:b}
  \end{subfigure}
  \caption{
  \textbf{BrickBench.}
  We visualize disaggregated BrickBench quantitative results.
  The baselines BrickNet-14B~\citep{Kulits_2026_CVPR} and BrickGPT~\citep{Pun_2025_ICCV} are only evaluated on the \texttt{Model} setting and the \texttt{Valid} metric reported of BrickNet-14B does not include stability.
  }
  \label{fig:metric-bars-t1}
\end{figure}

\begin{figure}[htb]
  \centering
  \captionsetup[subfigure]{skip=2pt}
  \begin{subfigure}{\textwidth}
    \centering
    \includegraphics[width=\linewidth]{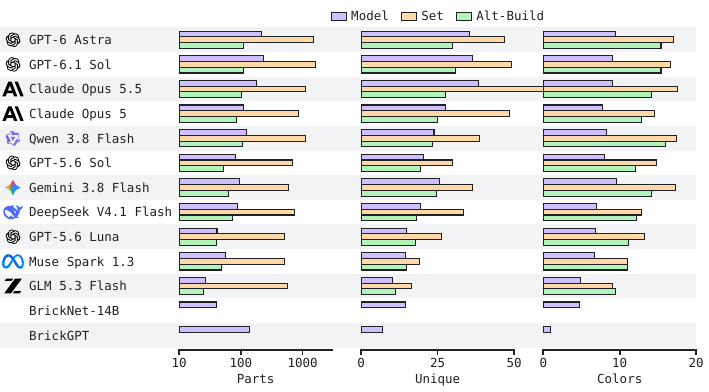}
    \subcaption{}\label{fig:metric-bars-t3:a}
  \end{subfigure}\\[4pt]
  \begin{subfigure}{\textwidth}
    \centering
    \includegraphics[width=\linewidth]{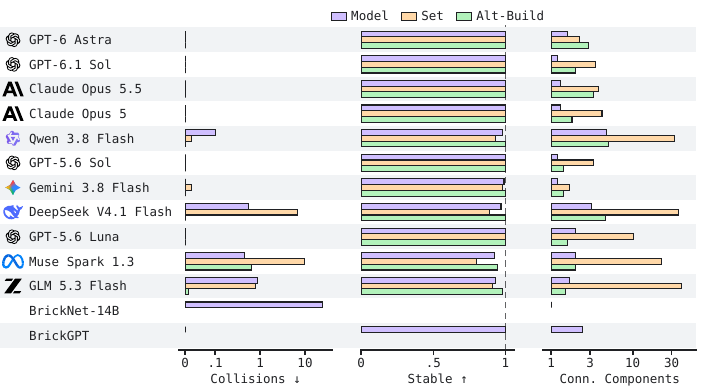}
    \subcaption{}\label{fig:metric-bars-t3:b}
  \end{subfigure}
  \caption{
  \textbf{Further assembly statistics.}
  We plot the statistics of \cref{tab:stats} by setting.
  }
  \label{fig:metric-bars-t3}
\end{figure}

\cref{fig:metric-bars-t1,fig:metric-bars-t3} plot \cref{tab:main,tab:stats} by setting in the layout of \cref{fig:metric-bars}, and \cref{fig:grid-appendix} shows one assembly per agent for further prompts in each setting.

\providecommand{\gridcol}[3]{%
  \makebox[\dimexpr\textwidth/11\relax][c]{%
    \raisebox{-0.5\height}{\includegraphics[height=11.5pt]{fig/icons/#1}}\hspace{1.5pt}%
    \raisebox{-0.5\height}{\sffamily\fontsize{6}{7}\selectfont
      \begin{tabular}{@{}l@{}}#2\\#3\end{tabular}}}}
\providecommand{\gridlabels}{%
  \vspace{3pt}\noindent%
  \gridcol{openai}{GPT-6}{Astra}\gridcol{openai}{GPT-6.1}{Sol}\gridcol{anthropic}{Claude}{Opus 5.5}%
  \gridcol{anthropic}{Claude}{Opus 5}\gridcol{qwen}{Qwen 3.8}{Flash}\gridcol{openai}{GPT-5.6}{Sol}%
  \gridcol{gemini}{Gemini 3.8}{Flash}\gridcol{deepseek}{DeepSeek}{V4.1 Flash}\gridcol{openai}{GPT-5.6}{Luna}%
  \gridcol{meta}{Muse}{Spark 1.3}\gridcol{zai}{GLM 5.3}{Flash}}

\begin{figure}[p]
  \renewcommand{\tsrpromptfont}{\fontsize{6.5}{7.4}\selectfont}%
  \centering
  \teaserrow{Model}{tsrModel}{A twisting gray tornado narrows to a thin point above a patch of churning water. Three small sharks protrude from different heights of the funnel, their bodies tilted in different directions. White bellies, fins, and open jaws break up the spiraling gray bands, beneath a broad cap of storm clouds.}{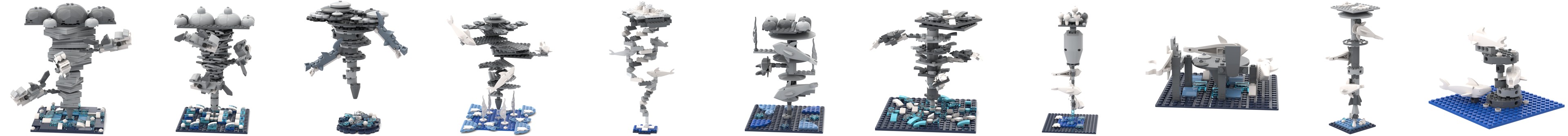}
  \teaserrow{Model}{tsrModel}{A saber-toothed squirrel is trapped inside a clear blue block of ice. Its long snout, enormous eyes, and curled bushy tail remain visible through the ice. One paw reaches out through a thawed opening toward a brown acorn lying just beyond its fingertips. A small puddle surrounds the bottom of the block.}{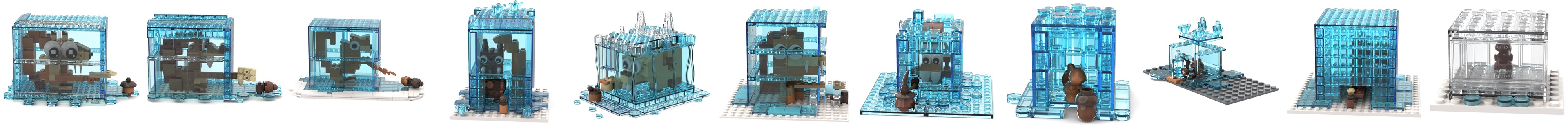}
  \teaserrow{Model}{tsrModel}{A battered red fox animatronic has an eyepatch, a long toothy snout, and one hand replaced by a metal hook. Torn fur exposes gray mechanical parts at its knees and chest. It wears ragged brown shorts, and its large metal feet stand apart. One ear is damaged, giving the already lopsided face an even more disreputable appearance.}{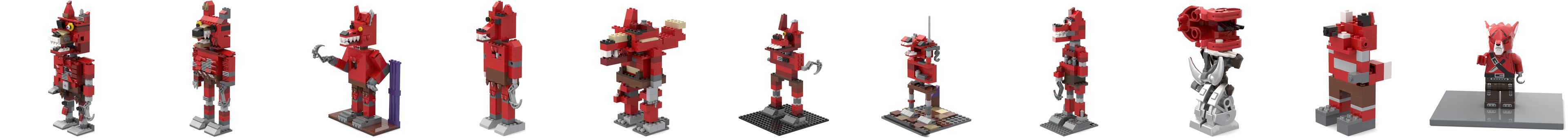}
  \teaserrow{Set}{tsrSet}{A cutaway medieval gatehouse contains a wooden catapult with a black-and-white cow sitting in its sling. A helmeted guard stands beside the release lever. Outside the closed gate, three knights look up from behind their shields. A second cow waits beside a hay bale in the courtyard.}{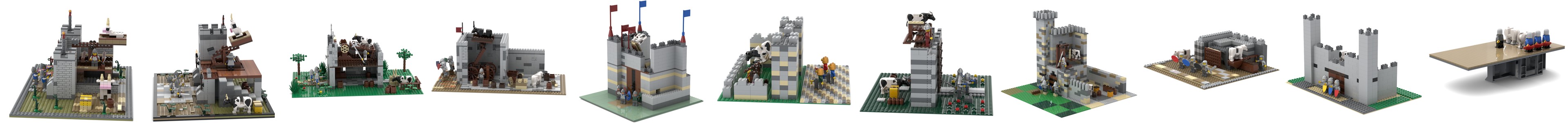}
  \teaserrow{Set}{tsrSet}{A rectangular glass fish tank with a black rim and colored gravel. Three orange fish and a striped angelfish swim among tall green plants and a stone arch. A snail clings to the inside glass, and a small filter occupies a corner across the tank from the stone arch.}{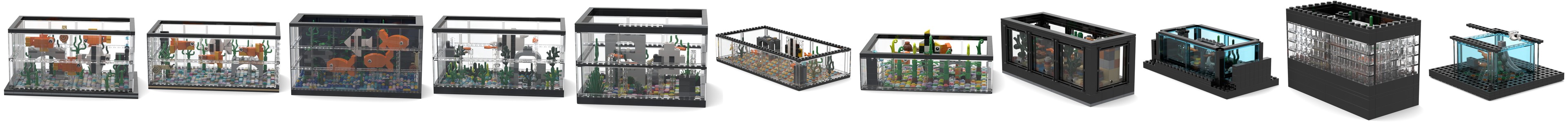}
  \teaserrow{Set}{tsrSet}{A barefoot martial artist wearing a white uniform, black belt, and red headband smashes a parked sedan on a waterfront platform. The car's hood is crumpled, its windows are broken, and one door hangs crookedly from the body. The fighter has one fist driven into the damaged front end. Harbor water and mooring posts stand behind the wreck.}{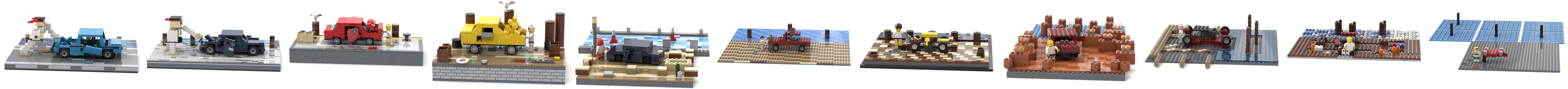}
  \teaserrow{Set}{tsrSet}{Four tiny children gather around a cream-filled cookie taller than they are in a jungle of enormous grass blades. Two sit on the back of a giant brown ant while another breaks a crumb from the cookie. The fourth peers into the opening of a discarded soda-can tab. Dewdrops cling to the grass above them.}{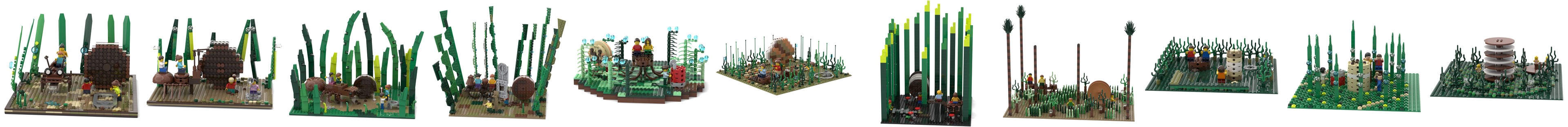}
  \teaserrow{Alt-Build}{tsrAlt}{A ferry carrying a car on its open deck. Its front ramp rests on the shore, and a small control cabin stands above the back of the deck.}{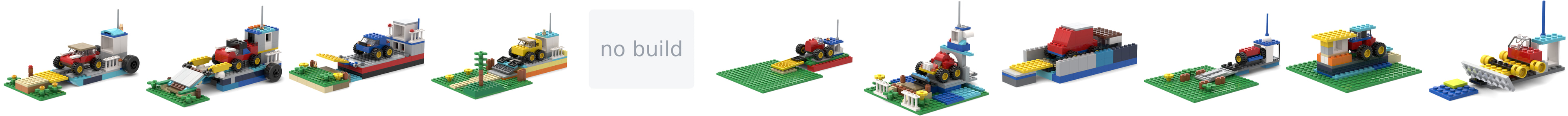}
  \teaserrow{Alt-Build}{tsrAlt}{A pink pig riding in a gray minecart. Its front legs rest over the rim, and its snout projects beyond the cart.}{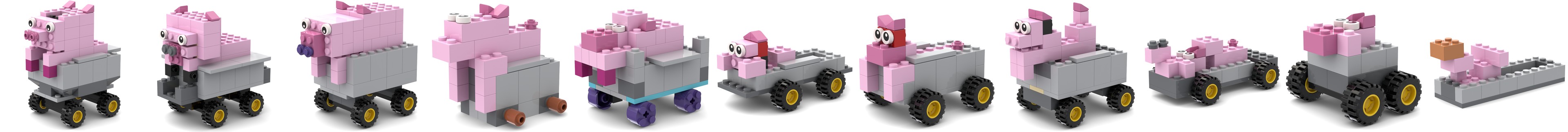}
  \teaserrow{Alt-Build}{tsrAlt}{A tree forks into two branches, with a nest between them. Two small birds sit in the nest. A larger bird perches on one branch.}{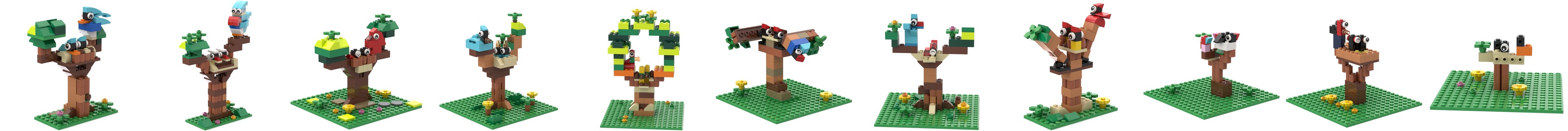}
  \gridlabels
  \caption{\textbf{Additional BrickBench samples.} We show further assemblies from the eleven agents in the three settings.}
  \label{fig:grid-appendix}
\end{figure}

\begin{figure}[htb]
  \centering
  \includegraphics[width=\linewidth]{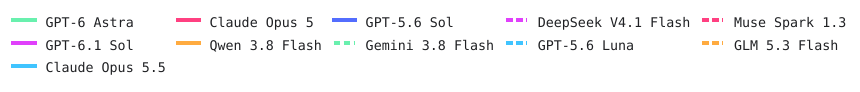}\\[3pt]
  \begin{subfigure}[t]{0.49\textwidth}
    \centering
    \includegraphics[width=\linewidth]{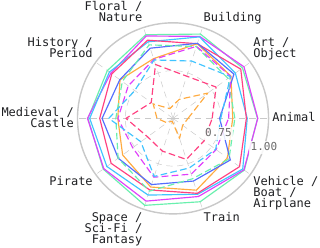}
    \caption{By prompt category}
    \label{fig:vqa-radar-cat}
  \end{subfigure}
  \hfill
  \begin{subfigure}[t]{0.49\textwidth}
    \centering
    \includegraphics[width=\linewidth]{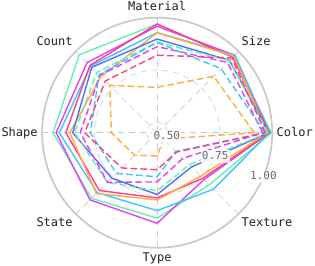}
    \caption{By attribute type}
    \label{fig:vqa-radar-attr}
  \end{subfigure}
  \caption{
    \textbf{VQA breakdowns.}
    We report score by prompt category (a) and attribute type (b).
}
  \label{fig:vqa-radars}
\end{figure}

\cref{fig:vqa-radars} breaks \texttt{VQA} down by build category and by attribute type.

\begin{figure}[htb]
  \centering
  \includegraphics[width=\linewidth]{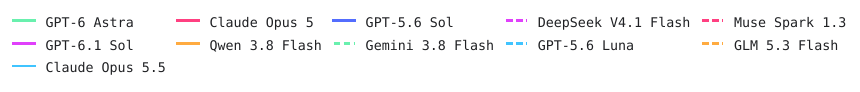}\\[3pt]
  \begin{subfigure}[t]{0.49\textwidth}
    \centering
    \includegraphics[width=\linewidth]{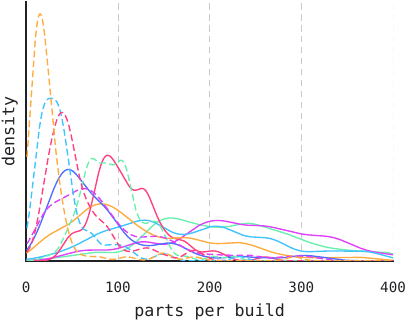}
    \caption{\texttt{Model}}
    \label{fig:parts-model}
  \end{subfigure}
  \hfill
  \begin{subfigure}[t]{0.49\textwidth}
    \centering
    \includegraphics[width=\linewidth]{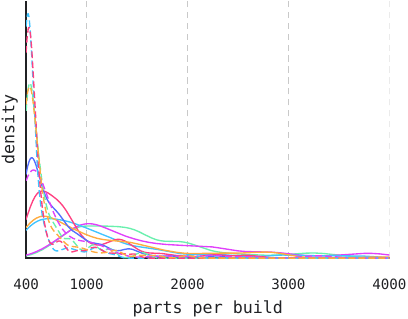}
    \caption{\texttt{Set}}
    \label{fig:parts-set}
  \end{subfigure}
  \caption{
    \textbf{Part-count distributions.}
    We report the density of part counts in the \texttt{Model} ($\leq$400) and \texttt{Set} (400--4000) settings for each agent.
    Most agents peak near 400 in \texttt{Set}.
    GPT-6 Astra uses considerably more parts on average in both settings.
}
  \label{fig:parts-density}
\end{figure}

\cref{fig:parts-density} plots the distribution of part counts in \texttt{Model} and \texttt{Set}.
Most agents peak near the 400-part floor of \texttt{Set}, and Astra is the exception.

\clearpage

\end{document}